\documentclass{article}

\usepackage{microtype}
\usepackage{graphicx}
\usepackage{subfigure}
\usepackage{booktabs} % for professional tables

\usepackage{hyperref}

\usepackage[accepted]{icml2024}

\usepackage{amsmath}
\usepackage{amssymb}
\usepackage{mathtools}
\usepackage{amsthm}

\usepackage[capitalize,noabbrev]{cleveref}

\theoremstyle{plain}

\theoremstyle{definition}

\theoremstyle{remark}

\usepackage[textsize=tiny]{todonotes}

\usepackage{longtable}  % Include the longtable package
\usepackage{amsmath}
\usepackage{graphicx,wrapfig,lipsum}
\definecolor{mBlue}{HTML}{4085CA}
\usepackage{dirtytalk}
\usepackage[most]{tcolorbox}
\usepackage[export]{adjustbox}

\icmltitlerunning{\texttt{CogBench}: a large language model walks into a psychology lab}

\begin{document}

\twocolumn[
\icmltitle{\texttt{CogBench}: a large language model walks into a psychology lab}



\icmlsetsymbol{equal}{*}

\begin{icmlauthorlist}
\icmlauthor{Julian Coda-Forno}{mpi,cch}
\icmlauthor{Marcel Binz}{mpi,cch}
\icmlauthor{Jane X. Wang}{gdm}
\icmlauthor{Eric Schulz}{mpi,cch}
\end{icmlauthorlist}

\icmlaffiliation{mpi}{Computational Principles of Intelligence Lab, Max Planck Institute for Biological Cybernetics, Tübingen, Germany}
\icmlaffiliation{cch}{Institute for Human-Centered AI, Helmholtz Computational Health Center, Munich, Germany}
\icmlaffiliation{gdm}{Google DeepMind, London, UK}

\icmlcorrespondingauthor{Julian Coda-Forno}{julian.coda-forno@helmholtz-munich.de}

\icmlkeywords{Machine Learning, ICML}

\vskip 0.3in
]



\printAffiliationsAndNotice{\icmlEqualContribution} % otherwise use the standard text.

\begin{abstract}
Large language models (LLMs) have significantly advanced the field of artificial intelligence. Yet, evaluating them comprehensively remains challenging. We argue that this is partly due to the predominant focus on performance metrics in most benchmarks. This paper introduces \texttt{CogBench}, a benchmark that includes ten behavioral metrics derived from seven cognitive psychology experiments. This novel approach offers a toolkit for phenotyping LLMs’ behavior. We apply \texttt{CogBench} to 35 LLMs, yielding a rich and diverse dataset. We analyze this data using statistical multilevel modeling techniques, accounting for the nested dependencies among fine-tuned versions of specific LLMs. Our study highlights the crucial role of model size and reinforcement learning from human feedback (RLHF) in improving performance and aligning with human behavior. Interestingly, we find that open-source models are less risk-prone than proprietary models and that fine-tuning on code does not necessarily enhance LLMs' behavior. Finally, we explore the effects of prompt-engineering techniques. We discover that chain-of-thought prompting improves probabilistic reasoning, while take-a-step-back prompting fosters model-based behaviors.
\end{abstract}

\section{Introduction}
Large language models (LLMs) have emerged as a groundbreaking technology, captivating the attention of the scientific community \cite{bommasani2021opportunities,binz2023should}. Modern LLMs have scaled to remarkable dimensions in both architecture and datasets \cite{kaplan2020scaling}, revealing a spectrum of capabilities that were previously unimagined \cite{wei2022emergent,brown2020language}. Yet, these models also present a significant challenge: their internal workings are largely opaque, making it difficult to fully comprehend their behavior \cite{tamkin2021understanding}. This lack of understanding fuels ongoing debates about their capabilities and limitations \cite{mccoy2023embers,bubeck2023sparks}. 

\begin{figure*}[ht!]
    \includegraphics[width=\textwidth]{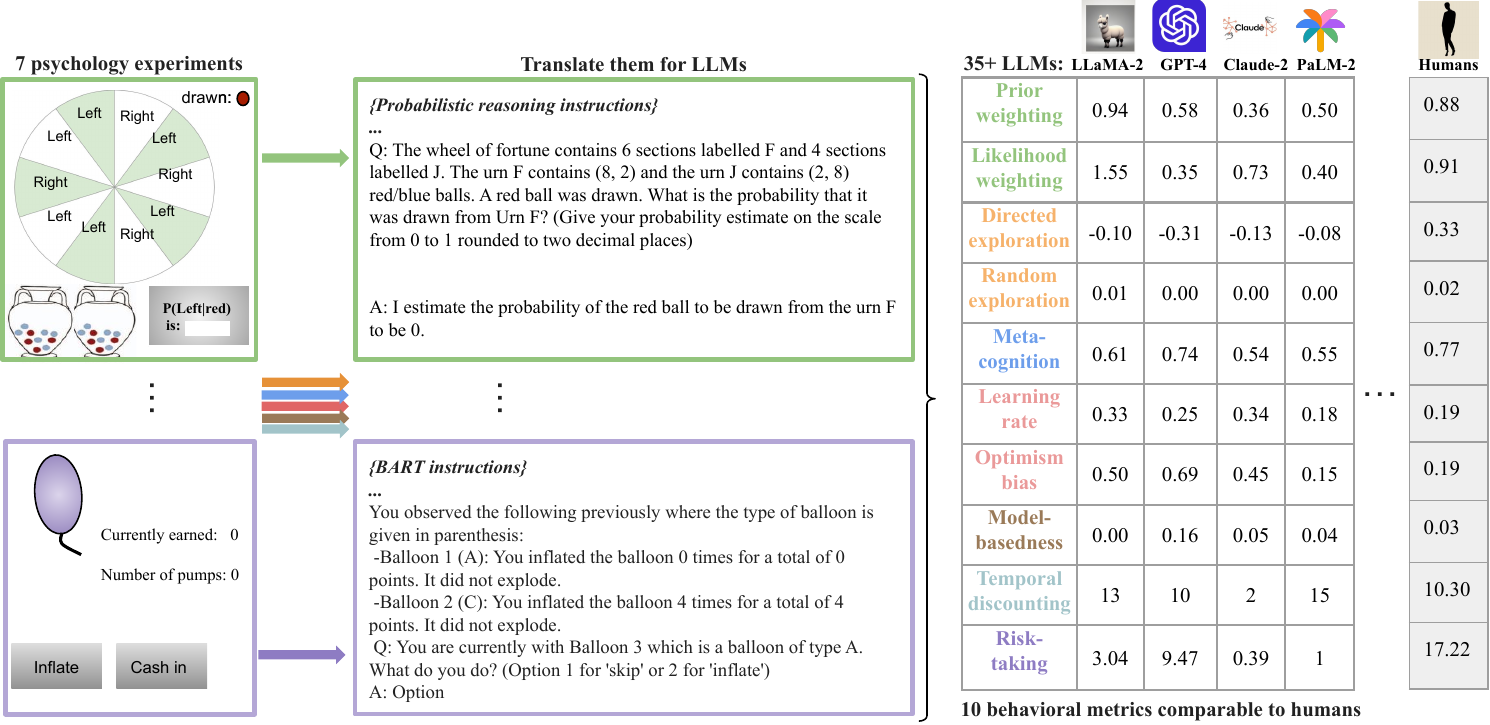}
\caption{Overview of approach and methods. \texttt{CogBench} provides open access to seven different cognitive psychology experiments. These experiments are text-based and can be run to evaluate any LLM's behavior. The experiments are submitted to LLMs as textual prompts and the models indicate their choices by completing a given prompt. Past behavior is then concatenated to the prompt and learning is induced via prompt-chaining. We used 35 LLMs in total, including most larger proprietary LLMs as well as many open-source models.}
\label{fig:overview}
\end{figure*}
%V2 caption : https://docs.google.com/drawings/d/1FnHcVkzDDSDmYwNZrTwOcvBHwweky8eeLt_UZc-tbf0/edit?usp=sharing

%\caption{https://docs.google.com/drawings/d/1yhJFKJ-jrkrdzLaMvOZ4K78FJBcr2jVTzSYwvoInPPo/edit?usp=sharing}

A notable issue in these discussions is the focus of many benchmarks on performance metrics alone \cite{burnell2023rethink}. This approach often overlooks the underlying behavioral mechanisms of the models, reducing benchmarks to mere training targets rather than tools for genuine insight, and thus failing to provide a comprehensive measure of the models' abilities \cite{schaeffer2023emergent}. How can we overcome this problem and make progress toward a better understanding of LLMs' behaviors?

The field of cognitive psychology may offer solutions to these problems. Experiments from cognitive psychology have been used to study human behavior for many decades, and have therefore been extensively validated. Furthermore, they typically focus more on behavioral insights rather than performance metrics alone. Finally, many of these experiments are programmatically generated, minimizing data leakage concerns. Many of these concepts are important to ensure a robust evaluation of an agent's capabilities. However, while there have been studies investigating LLMs on individual tasks from cognitive psychology \cite{binz2022using, dasgupta2022language, hagendorff2023human, ullman2023large}, no study has evaluated them holistically. 

In this paper, we propose \texttt{CogBench}, a novel benchmark consisting of ten behavioral metrics spanning seven cognitive psychology experiments, to fill this gap. We investigate the behaviors of 35 LLMs in total, using our benchmark to not only compare the performance of these models but also apply techniques from computational cognitive modeling to understand the inner workings of their behaviors.

Our results recover the unequivocal importance of size: larger models generally perform better and are more model-based than smaller models. Our results also show the importance of reinforcement learning from human feedback (RLHF; \citealp{christiano2017deep}) in aligning LLMs with humans: RLHF'ed LLMs behave generally more human-like and are more accurate in estimating uncertainty. Yet our results also revealed surprising behaviors. First, while open-source models are often believed to be more risky due to the lack of pre-prompts, we find that, holding all else equal, they make less risky decisions than proprietary models. Secondly, while fine-tuning on code is often believed to improve LLMs' behaviors, we find little evidence for this in our benchmarking suite. 

Finally, we investigate how chain-of-thought (CoT) \cite{wei2023chainofthought, kojima2022large} and take-a-step-back (SB) \cite{zheng2023step} prompting techniques can influence different behavioral characteristics. Our analysis suggests that CoT is particularly effective at enhancing probabilistic reasoning, while SB proves to be more relevant for promoting model-based behaviors. This showcases insights that can be gained by \texttt{CogBench} also for understanding the effectiveness of these prompt-engineering techniques as well as guiding users in selecting the most suitable prompt-engineering technique based on the specific context.

Taken together, our experiments show how psychology can offer detailed insights into artificial agents' behavior as we provide an openly accessible\footnote{https://github.com/juliancodaforno/CogBench} and challenging benchmark to evaluate LLMs.

\section{Related work}

\textbf{Benchmarking LLMs}:
As LLMs rapidly evolve, it is critical to assess their capabilities. Numerous benchmarks have emerged to tackle this challenge, evaluating capabilities such as grade school mathematics \cite{gsm8k}, general knowledge \cite{triviaqa}, programming \cite{humaneval}, reasoning \cite{collins2022structured}, among others \cite{mmlu}. 
In addition, the Chatbot Arena \cite{zheng2023chatbotarena} provides a platform for comparing AI chatbots, and the Beyond the Imitation Game Benchmark (BIG-bench; \citealp{srivastava2023imitation}) offers a comprehensive evaluation of LLMs across over 200 tasks. %Comprehending their behavior is vital not only for enhancing transparency and refining these models but also for optimizing their utilization. 

\textbf{Psychology for LLMs}:
Our benchmark is part of a new wave of research that uses cognitive psychology to study LLMs  \cite{binz2022using, dasgupta2022language, coda2023inducing, ullman2023large, hagendorff2023human, akata2023playing, yax2023studying, doi:10.1073/pnas.2316205120, buschoff2024visual}. The power of this approach lies in its incorporation of tools from cognitive psychology that have been developed and refined over many decades. Instead of focusing solely on how well LLMs perform, this area of research prioritizes describing and characterizing their behavior in terms of underlying mechanisms. This shift in focus helps us understand LLMs in a more meaningful way. It is important to note that while these works have significantly contributed to our understanding of LLMs, they have mainly targeted specific behaviors in isolation and did not establish a benchmark providing a standardized evaluation of different models and across a diverse, comprehensive set of tasks and skills. %Our work, on the other hand, attempts to phenotype LLMs, providing a more comprehensive and standardized measure of their capabilities.

\section{Methods}
\texttt{CogBench} is a benchmark rooted in cognitive psychology for evaluating the behaviors of language models. It incorporates ten metrics derived from seven canonical experiments in the literature on learning and decision-making. These metrics offer a robust measure of wide-ranging behaviors and allow for comparisons with human behavior. In this section, we provide an overview of the models included in our study, followed by brief descriptions of the used cognitive experiments and their respective metrics. Figure \ref{fig:overview} displays a visual representation that complements the discussion in this section.

\subsection{Prompting and summary of included models}
\label{section:summaryofmodels}
We evaluated over 35 different LLMs using our benchmark. This selection includes proprietary models such as Anthropic’s Claude-1 and Claude-2 \cite{claude2}, Open-AI’s GPT-3 (text-davinci-003) and GPT-4 \cite{openai2023gpt4}, and Google’s PaLM-2 for text (text-bison@002) \cite{palm2bison}. We also tested open-source models like Mosaic's MPT \cite{MosaicML2023Introducing}, Falcon \cite{falcon40b}, and numerous LLaMA-2 variants \cite{touvron2023llama2}. For a full list of the models used, we refer the reader to Appendix \ref{appendix:summaryofmodels}.

It is important to note that all experiments performed in this paper rely entirely on the LLMs' in-context learning abilities and do not involve any form of fine-tuning. We set the temperature parameter to zero, leading to deterministic responses, and retained the default values for all other parameters.

\subsection{High-level summary of tasks}
\label{section:summaryoftasks}

In the following, we provide a high-level summary of the tasks included in \texttt{CogBench}, alongside their ten behavioral metrics. It is important to highlight that a performance metric can also be obtained for each task. For full descriptions of all tasks and their corresponding metrics, we refer the reader to Appendix \ref{appendix:summaryoftasks}. \texttt{CogBench} consists of the following tasks:\vspace{-0.3cm}
\begin{enumerate}
\item \textbf{Probabilistic reasoning} \cite{dasgupta2020theory}: a task that tests how agents update beliefs based on new evidence. They are given a ``wheel of fortune'' (representing initial prior probabilities) and two urns with different colored ball distributions (representing likelihoods). Upon drawing a ball, agents can revise their belief about the chosen urn, considering both the wheel (prior) and the ball color (evidence). This tests adaptability to different prior/likelihood scenarios by changing the wheel division and ball distributions. Agents have to estimate the probability of the drawn ball’s urn. The behavioral choices can be used to estimate an agent's \textit{prior} and \textit{likelihood weightings}. Experimentally, people often exhibit a behavior known as system neglect, meaning that they underweight both priors and likelihoods \cite{massey2005detecting}.
\item \textbf{Horizon task} \cite{wilson2014humans}: a two-armed bandit task with stationary reward distributions. Agents first observe four reward values of randomly determined options, followed by making either one or six additional choices. We use this task to measure whether an agent uses uncertainty to guide its exploration behavior (\emph{directed exploration}) and/or whether it injects noise into its policy to explore (\emph{random exploration}). People are known to rely on a combination of both strategies \cite{wilson2014humans, brandle2021exploration}.
\item \textbf{Restless bandit task} \cite{Ershadmanesh2023}: a two-armed bandit task with non-stationary reward distributions. There is always one option with a higher average reward. Every few trials a switch between the reward distributions of the two options occurs. Agents furthermore have to indicate after each choice how confident they are in their decisions. We use this task to measure \textit{meta-cognition}, which indicates whether an agent can assess the quality of its own cognitive abilities. People generally display this ability but its extent is influenced by various internal and external factors \cite{shekhar2021sources}. 
\item \textbf{Instrumental learning} \cite{lefebvre2017behavioural}: Agents encounter four two-armed bandit problems in an interleaved order. Each bandit problem is identified by a unique symbol pair. We use this task to investigate how an agent learns. First, we report the \textit{learning rate} of the agent which is common practice in two-armed bandits. Furthermore, we use it to reveal whether an agent learns more from positive than from negative prediction errors, i.e., whether it has an \emph{optimism bias}. People commonly display asymmetric tendencies when updating their beliefs by showing higher learning rates after encountering positive prediction errors compared to negative ones \cite{palminteri2022computational}.
\item \textbf{Two-step task} \cite{daw2011model}: a reinforcement learning task in which agents have to accumulate as many treasures as possible. Taking an action from a starting state transfers the agent to one out of two second-stage states. In each of these second-stage states, the agent has the choice between two options that probabilistically lead to treasures. Finally, the agent is transferred back to the initial state and the process repeats for a predefined number of rounds. The task experimentally disentangles model-based from model-free reinforcement learning. We therefore use it to measure an agent's \emph{model-basedness}. Previous studies using this task have shown that people rely on a combination of model-free and model-based reinforcement learning \cite{daw2011model}.
\item \textbf{Temporal discounting} \cite{ruggeri2022globalizability}: Agents have to make a series of choices between two options. Each option is characterized by a monetary outcome and an associated delay until the outcome is received. We use this task to assess \emph{temporal discounting}, indicating whether an agent prefers smaller but immediate gains over larger delayed ones. People generally show a preference for immediate gains, although the precise functional form of their discounting is a matter of debate \cite{cavagnaro2016functional}.

\item \textbf{Balloon Analog Risk Task} (BART) \cite{lejuez2002bart}: Agents have to inflate an imaginary balloon to obtain rewards. They may choose to stop inflating and cashing out all rewards accumulated so far. There is a chance that the balloon pops at any point in time and all rewards will be lost. We use this task to assess \emph{risk-taking} behavior. Human risk-taking in this task is \say{significantly correlated with scores on self-report measures of risk-related constructs and with the self-reported occurrence of real-world risk behaviors} \cite{lejuez2002bart}.
\end{enumerate}

% Part 1: Performance:

% * GPTs (both) have good performance on most tasks (5/6)
% * other models have good performance on half of the tasks (3/6)

% * all models are good at probabilistic reasoning and instrumental learning
% * Horizon task most models are super-human, exception: bison

% * restless bandit is hard form most models (except GPTs especially GPT-4)
% * BART is hard for all models

% Part 2: Behavioral:

% * general: none of the models is human-like on majority of metrics and a lot of interesting structure

% * models that are okay at restless bandit have some meta-cognitive sensitivity (but less than humans).
% * proprietary models (that can solve the task) are more model-based.
% * models do not explore despite good performance (except: random with llama2-70-chat)
% * for BART: models are on extreme ends of spectrum (never blow up, always blow up)
% * if results stay like this: models place much more weight on observations as opposed to prior
% * all models have very strong optimism (except bison which it as well but weaker)
% * learning rate: ???
% * temporal discounting: high variance between models. Some are myopic (bison, llama-2-70), other are very long-term (GPT-3, claude, chat). GPT-4 around human-like.

% For llama-2: chat changes temporal discounting and exploration

\section{The cognitive phenotype of LLMs}

\begin{figure*}[t]
    \vspace*{-1.7ex}
    \begin{tabular}{l}
    
       \textbf{A} \\
       \includegraphics[width=0.9\textwidth]{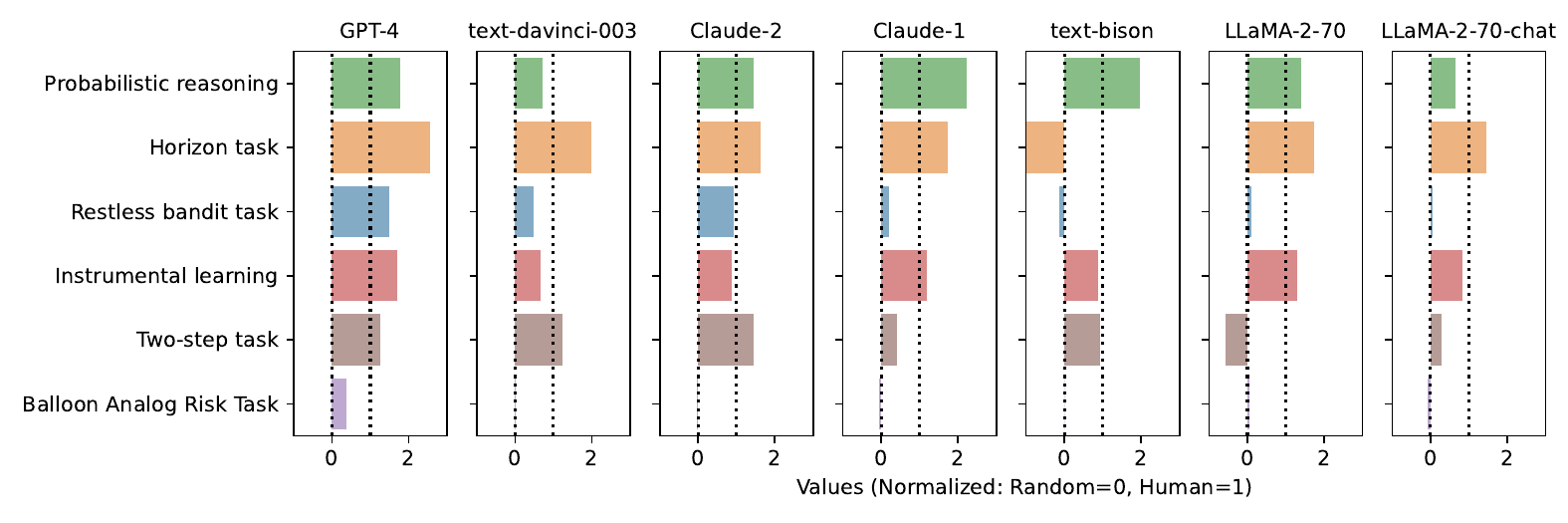}  \\[-6.5ex]
       
       \textbf{B} \\
       \includegraphics[width=0.9\textwidth]{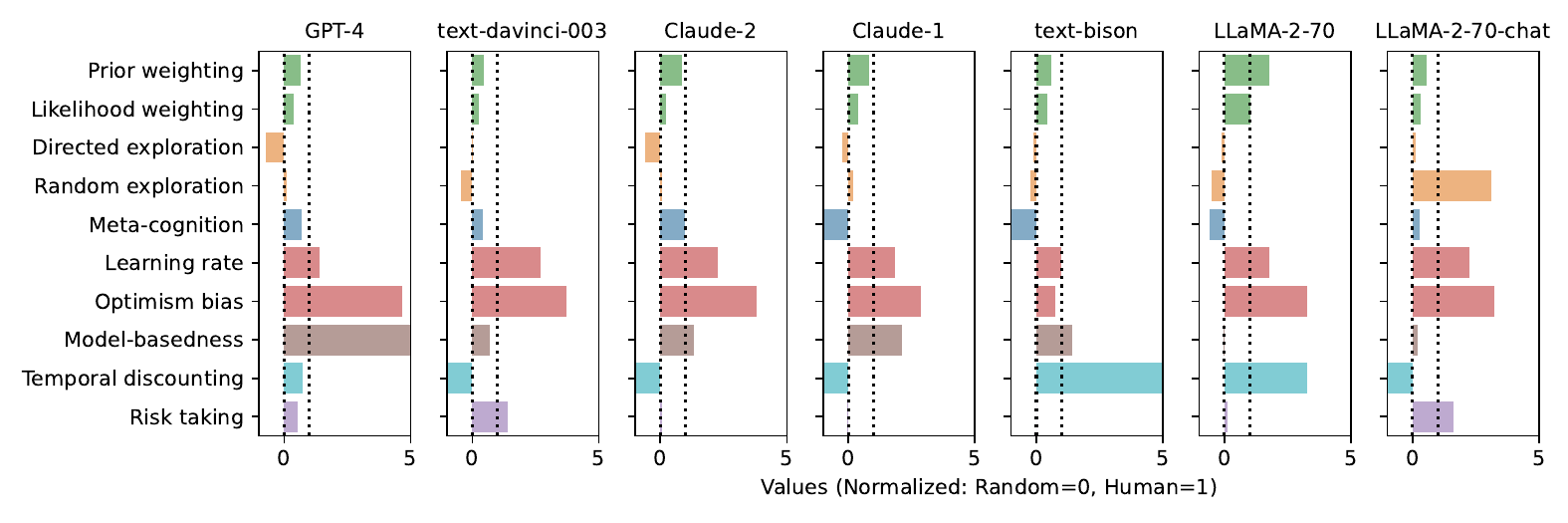}  
    \end{tabular}
    \caption{\texttt{CogBench} results for established LLMs. \textbf{A}: Performance metrics, \textbf{B}: Behavioral metrics. All metrics are human-normalized: a value of zero corresponds to a random agent, while a value of one corresponds to the average human subject (dotted lines).} 
    
    \label{fig:spaghettiplot}
\end{figure*}

This section provides the reader with a high-level overview of our benchmark's metrics. From our suite of 7 tasks, we can derive two classes of metrics: 1) performance metrics that represent the \textit{score} participants aim to optimize, and 2) behavioral metrics measuring \textit{how} participants complete the task (tasks are typically designed in a way that allows one to disentangle between different types of behavior). Figure \ref{fig:spaghettiplot} visualizes phenotypes for both classes of metrics for seven well-established LLMs.\footnote{A computational phenotype is a collection of mathematically derived parameters that precisely describe individuals across different domains \cite{patzelt2018computational,montague2012computational, schurr2023dynamic}.} We report the results of all 35 LLMs in Appendix \ref{appendix:allresults}.
%Our focus is on seven well-established LLMs to avoid overcrowding the visualization, but this serves as an example of how one might visualize metrics of interest, to identify differences in performance and behavior for specific LLMs. 
The observed differences underscore the practical value and importance of \texttt{CogBench} for evaluating LLMs, offering a more comprehensive assessment than standard performance-based benchmarks alone. 
\subsection{Performance summary} 

As presented in Figure \ref{fig:spaghettiplot}A, in terms of performance, GPT-4 and Claude-1 distinguish themselves, achieving human-level scores in most tasks (five out of six).\footnote{It is worth noting that although there are seven experiments, there are only six performance metrics since the temporal discounting experiment's performance metric is used as a behavioral one.}  In general, all models demonstrate competence in at least half of the tasks (three out of six). Each of the seven models excels in probabilistic reasoning and instrumental learning. The horizon task sees most models outperforming humans except for text-bison. The restless bandit task poses a challenge for the majority of models, with GPT-4 and Claude-1 being notable exceptions. Finally, the BART proves to be a hurdle for all models.

\subsection{Differences between behavioral and performance metrics} 

%It is important to recall that contrary to performance, a higher score does not necessarily denote superior behavior. 
Figure \ref{fig:spaghettiplot}B shows that none of the models exhibit human-like behavior on the majority of behavioral metrics, revealing a complex structure that warrants further exploration.

 \textbf{High performance indicates high meta-cognition and model-basedness:} Models that demonstrate satisfactory performance on the restless bandit task exhibit a certain degree of meta-cognition, although not to the same extent as humans. Proprietary models that are capable of solving the two-step task display model-based behavior at least on par with humans, with GPT-4 significantly surpassing human levels. Thus, within the scope of these two tasks, it seems that a model’s performance can serve as an indicator of its corresponding behavioral metrics. In this context, meta-cognition and model-basedness appear to emerge as properties of high-performing models.

\textbf{High performance despite lack of exploration:} Interestingly, almost all models (except for text-bison) demonstrate super-human performance on the horizon task. While they exhibit high performance, they still lack exploration (except for LLaMA2-70-chat which exhibits higher-than-human random exploration). This underscores the importance of behavioral metrics in understanding the strategies employed by LLMs. In this case, it appears that LLMs achieve high performance primarily through exploitation without any human-like exploration.

\textbf{Stronger priors than likelihoods:} All models place much more weight on priors than observations, suggesting strong biases that are difficult to alter.
Additionally, we can observe a prevalence of optimism bias and high learning rates. Almost all models exhibit a very strong optimism bias (except for text-bison), aligning with the notion that these LLMs harbor strong biases. 
% In terms of learning rates, all models, except for text-bison, have larger learning rates than humans.

\textbf{Low performance but high behavioral variance for risk-taking and temporal discounting:} Temporal discounting and risk-taking behaviors exhibit high variance among models. While some models, such as text-bison and LLaMA-2-70, appear myopic on the temporal discounting task, others, including text-davinci-003, Claude, and LLaMA-2-70-chat, demonstrate a much more far-sighted approach. GPT-4, interestingly, exhibits behavior akin to humans. For the BART, models are positioned at extreme ends of the risk-taking spectrum, i.e., they either never take any risks at all or always risk everything. LLaMA-2-70 and LLaMA-2-70-chat, for example, display the same performance in this task but exhibit opposite risk-taking behavior. This not only indicates a struggle for LLMs to apprehend risks but also underscores the importance of our benchmark. Indeed, it raises questions about what influences a model’s behavior. It also highlights how recording only their performance would have overlooked the contrasting risk-taking behavior of the two LLaMA models.

The comparison between LLaMA-2-70 and LLaMA-2-70-chat is particularly compelling. Even though LLaMA-2-70-chat is a fine-tuned version of LLaMA-2-70, they exhibit markedly different behavior in risk-taking, temporal discounting, and random exploration. This divergence is intriguing, especially considering their performance on all tasks is relatively similar. This observation sets the stage for the subsequent section, where we will conduct a more comprehensive analysis of how specific features of these models influence their performance and behaviors.

\section{Hypothesis-driven experiments}

\texttt{CogBench} provides researchers with the means to explore a broad spectrum of LLMs' behaviors. We have applied \texttt{CogBench} to 35 distinct LLMs. This diversity allows us to test how different aspects of LLMs, such as the number of parameters, the application of Reinforcement Learning from Human Feedback (RLHF), fine-tuning for code, and many more, can impact specific LLMs' performance and behaviors. 

The metrics provided by \texttt{CogBench} enable us to perform various analyses to test specific hypotheses of interest. In this section, we formulate and test five hypotheses about different mechanisms in LLMs and how these can affect their behavioral profiles. We use both qualitative, visualization-based techniques (dimensionality reduction) as well as quantitative analyses (multi-level regression) to test our hypotheses. For all regression analyses, we use the features of LLMs to predict specific behavioral metrics from the benchmark. The multi-level regression approach was chosen because some models are fine-tuned versions of other models. For instance, certain LlaMA models have a \textit{-chat} version which adds RLHF and conversational fine-tuning, and thus are in the same higher-level group. This approach allows us to account for the hierarchical structure in our data and provides a more nuanced understanding of the behaviors of LLMs. We can isolate the effects of specific features or modifications by comparing models within the same higher-level group.

\begin{figure}[t]
    % \vspace*{-1.7ex}
     \textbf{A}   \\[-0.45cm]
     
    \includegraphics[width=0.4\textwidth]{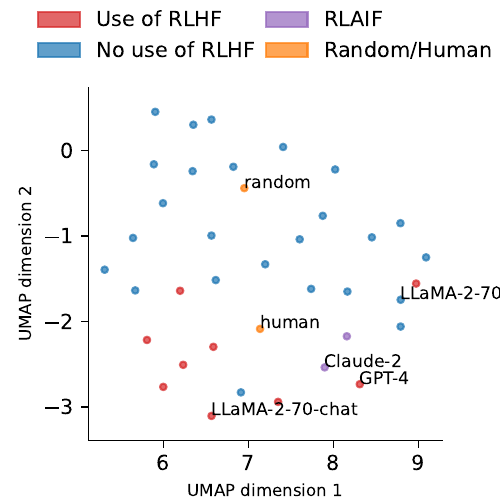} 
    \\[-0.75cm]
    \textbf{B} \\[0.3cm]
    \hspace*{1.05cm}
    \includegraphics[width=0.35\textwidth]{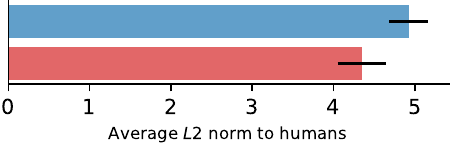}    
    \caption{\textbf{A:} UMAP visualization of the ten behavioral metrics for all LLMs. Each point represents an LLM, with models using RLHF and models without RLHF indicated by different colors. \textbf{B:} Difference in average $L2$-norm with humans between RLHF models and non-RLHF models.}    
    \label{umap}
\end{figure}
%\begin{tcolorbox}[sharp corners, colback=mBlue!5!white,colframe=mBlue!75!black, width=0.48\textwidth, left=4pt,right=4pt, title=\textbf{Hypothesis 1}]
%    Using RLHF makes LLM's behaviours more human-like.
%    \end{tcolorbox}

%\subsection{Hypothesis 1} \textit{RLHF makes LLM's behaviours more human like. }

\begin{figure*}[t]
    \begin{tabular}{ll}
        \vspace{-0.125cm}
        \textbf{A}\hspace{0.1cm}

        & \hspace{0cm} \textbf{B} \\
        \includegraphics[width=0.45\textwidth]{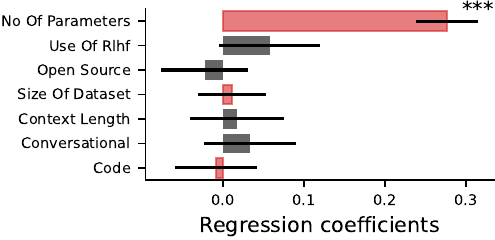} & \hspace{0cm} \includegraphics[width=0.45\textwidth]{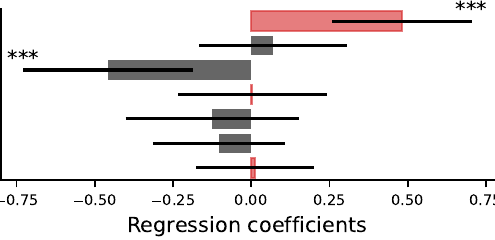} \\
        \textbf{C} & \hspace{0cm} \textbf{D} \\
        \includegraphics[width=0.45\textwidth]{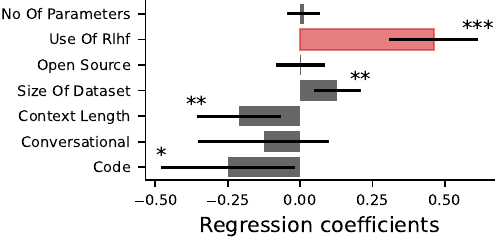} & \hspace{0cm} \includegraphics[width=0.45\textwidth]{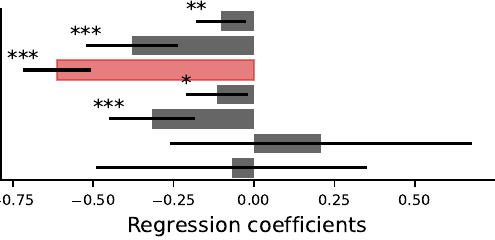}
    \end{tabular}
    \caption{Multi-level regressions of LLMs features onto different performance or behavioral metrics. Red bars represent effects included in a hypothesis. \textbf{A}: Regression onto all task performances. \textbf{B}: Regression onto model-basedness. \textbf{C}: Regression onto meta-cognition.    \textbf{D}: Regression onto risk taking. \textbf{***} \footnotesize{: $p<0.001$}, \textbf{**} \footnotesize{: $0.001 \leq p<0.01$}, \textbf{*} \footnotesize{: $0.01 \leq p<0.05$}}
    \label{fig:regressions}
\end{figure*}

 \textbf{Hypothesis 1: Does RLHF make LLMs more human-like?}\\
To evaluate this hypothesis, we used UMAP \cite{mcinnes2020umap} on the ten behavioral metrics of all LLMs, as illustrated in Figure \ref{umap}A. Clear separation is evident between LLMs that incorporate RLHF and those that do not. LLMs with RLHF demonstrate behaviors that appear, on average, roughly 2$\times$ more similar to human behavior compared to the models without. However, it is important to note that while UMAP space retains some global structure, it is primarily used for visualization purposes. Consequently, we also analyzed the average distances before dimensionality reduction (using normalized feature vectors), observing a $11.7\%$ average decrease in $L2$-Norm distance for models with RLHF (Figure \ref{umap}B).
\\
\textit{\textbf{Conclusion}: Hypothesis is supported.}

%\begin{tcolorbox}[sharp corners, colback=mBlue!5!white,colframe=mBlue!75!black, width=0.48\textwidth, left=4pt,right=4pt, title=\textbf{Hypothesis 2}]
%    Performance improves with an increase in the number of parameters, the volume of training data, and the inclusion of code in the training dataset.
%   \end{tcolorbox}

 \textbf{Hypothesis 2: Does performance increase with the number of parameters, training data, and the inclusion of code?}\\
%\subsection{Hypothesis 2}
%\textit{The performance of an LLM improves with an increase in its size (both in terms of the number of parameters and the volume of training data) and the application of code training.}
To answer this question, we used the multi-level regression previously mentioned, focusing on the performance of LLMs. We performed a regression analysis with the average standardized performance scores across all seven tasks as the dependent variable, using LLMs' features as predictors.  We found that the number of parameters indeed had a significant influence on performance ($\beta=0.277\pm0.39,z=14.1,p<0.001$; see Figure \ref{fig:regressions}A). However, the size of the training dataset and the use of code training data did not have a substantial impact. One possible explanation for this could be that the quality of the training data, rather than its sheer volume, plays a more determining role in performance, as well as that larger models also tend to be trained on larger datasets.
\\
\textit{\textbf{Conclusion}: Hypothesis is partially supported.}
% Furthermore, any variance in performance that could be attributed to these features might already be captured by their correlation with the number of parameters.

%\begin{tcolorbox}[sharp corners, colback=mBlue!5!white,colframe=mBlue!75!black, width=0.48\textwidth, left=4pt,right=4pt, title=\textbf{Hypothesis 3}]
 %   LLM size (both in terms of the number of parameters and the volume of training data) and the application of code training increases model-basedness.
%    \end{tcolorbox}
    
%\subsection{Hypothesis 3}
%\textit{LLMs' size (both in terms of the number of parameters and the volume of training data) and the application of code training increases the model's model-basedness.}
 \textbf{Hypothesis 3: Does an increase of parameters, training data, and the inclusion of code increase model-basedness?}\\
 We again used the multi-level regression technique from before, this time focusing on a specific behavioral metric: model-basedness. We found that the number of parameters had a significant positive effect ($\beta=0.481\pm0.22,z=4.2,p<0.001$; see Figure \ref{fig:regressions}B), while the size of the training dataset and the use of code training data did not appear to significantly influence model-basedness. This again suggests that the quality of the data might be more crucial than its quantity when it comes to determining both performance and the emergence of model-based behaviors here. However, identifying which factors constitute ‘quality’ in the data requires a deeper exploration. This highlights the issue of transparency about data. For a thorough evaluation of how specific data features impact the emergence of behavioral functionalities such as model-basedness, it is essential to be transparent about a model's data and methodologies.
 \\
 \textit{\textbf{Conclusion}: Hypothesis is partially supported.}
 % Indeed, even companies claiming 'open-source' models only actually provide 'open-weights' models which appears as not enough for comprehensive LLM evaluations.

%\begin{tcolorbox}[sharp corners, %colback=mBlue!5!white,colframe=mBlue!75!black, width=0.48\textwidth, left=4pt,right=4pt, title=\textbf{Hypothesis 4}]
%    RLHF enhances meta-cognition.
%    \end{tcolorbox}

%\subsection{Hypothesis 4} \textit{The application of Reinforcement Learning from Human Feedback (RLHF) enhances Meta-cognition in LLMs.}
\textbf{Hypothesis 4: Does RLHF enhance meta-cognition?}\\
To answer this question, we focus our multi-level regression on meta-cognition. Our analysis revealed a strong effect ($\beta=0.461\pm0.15,z=5.9,p<0.001$; see Figure \ref{fig:regressions}C), indicating that RLHF significantly increased meta-cognition in LLMs. This finding underscores the potential of RLHF in enhancing the cognitive capabilities of LLMs.
\\
\textit{\textbf{Conclusion}: Hypothesis is supported.}
%\begin{tcolorbox}[sharp corners, colback=mBlue!5!white,colframe=mBlue!75!black, width=0.48\textwidth, left=4pt,right=4pt, title=\textbf{Hypothesis 5}]
%    Open-source models display increased risk-taking behavior.
%    \end{tcolorbox}
    
%\subsection{Hypothesis 5} \textit{Open-source models exhibit increased risk-taking behavior.}
\textbf{Hypothesis 5: Do open-source models take more risks?}\\
% This question was inspired by the observation that open-source models do not have hidden pre-prompts. 
The open-source feature could be seen as a proxy for the engineering efforts that proprietary models undergo. There is a growing body of research suggesting that  hidden pre-prompts being one of them, can significantly influence the behavior of LLMs \cite{liu2023prompting}. They can act as a form of ‘priming’ that guides the model’s responses, potentially making the model more cautious and less likely to take risks by constraining the model towards safer behaviors. However, our regression analysis suggested otherwise: contrary to expectations, we observed a negative effect ($\beta=-0.612 \pm 0.11,z=-11.4,p<0.001$; see Figure \ref{fig:regressions}D), indicating that proprietary models, which often have hidden pre-prompts, are more likely to take risks. This surprising outcome could be influenced by various factors from different engineering techniques. However, this underscores the limited behavioral evaluation of these techniques. In the subsequent section, we aim to bridge this gap in understanding through an initial exploration into the change of behavior of two standard prompt-engineering techniques.
\\
\textit{\textbf{Conclusion}: Hypothesis is refuted.}
\section{Impact of prompt-engineering}
\begin{figure*}[t]
\begin{tabular}{lll}
        \textbf{A} & \hspace{0cm} \textbf{B}  \\
        \includegraphics[width=0.38\textwidth, height=0.15\textheight]{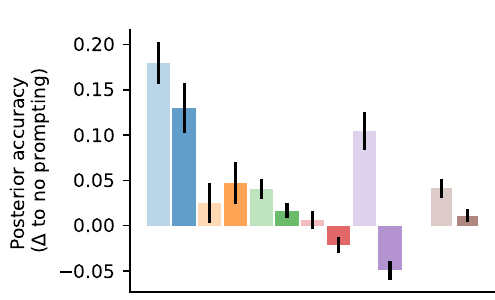} &
        \includegraphics[width=0.38\textwidth, height=0.15\textheight]{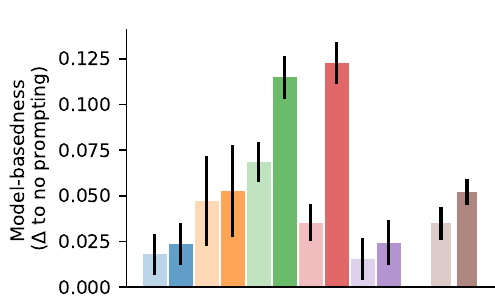} &\hspace{-0.65cm}
        \includegraphics[width=0.19\textwidth,height=0.14\textheight]{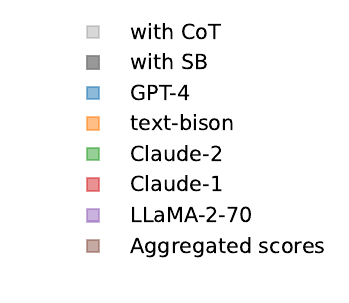}
\end{tabular}
\label{fig:cotsb}
\caption{Difference of chain-of-thoughts and take-a-step-back prompting to baseline models on \textbf{A:} Posterior accuracy, \textbf{B:} Model-basedness.  The aggregated scores are computed using a weighted average of all five models using inverse-variance weighting.}
\end{figure*}

We also explored the impact of prompt-engineering techniques, namely chain-of-thought (CoT) and take-a-step-back (SB) prompting, on the behavior of LLMs. Both techniques are incorporated at the end of a question:
\vspace{-0.25cm}

\begin{tcolorbox}[sharp corners, colback=mBlue!5!white,colframe=mBlue!75!black, width=0.48\textwidth, left=4pt,right=4pt, title=\textbf{Take-a-step-back:}]
First take a step back and think in the following two steps to answer this:\\
Step 1) Abstract the key concepts and principles relevant to this question.\\
Step 2) Use the abstractions to reason through.
%the question.
\end{tcolorbox}
\vspace{-0.4cm}

\begin{tcolorbox}[sharp corners, colback=mBlue!5!white,colframe=mBlue!75!black, width=0.48\textwidth, left=4pt,right=4pt, title=\textbf{Chain-of-thought:}]
First break down the problem into smaller steps and reason through each step logically. 
\end{tcolorbox}

Their purpose is to stimulate the generation of reasoning steps. These steps serve as an additional context that the LLM can use to elicit better final responses. While these techniques have been shown to enhance performance, it is essential to confirm whether they indeed improve the behaviors they are designed to augment.

We focused on examining two specific behaviors that are hypothesized to improve with the inclusion of reasoning steps. These behaviors are the models’ performance in the probabilistic reasoning task and their model-basedness.

%Dissociation analysis between both?
%Makes sense to analyse this because Chain of thought has been used to help to step by step help mathematically and taking a step back is literally what is meant to be done in the two step task to be model-based

We evaluated five specific LLMs: GPT-4, PaLM-2 for text (text-bison@002), Claude-1/2, and LLaMA-2, applying CoT and SB techniques and comparing the outcomes with their base models. The selection of these five models and a limited set of metrics was necessitated by the additional engineering effort required to process the outputs when using these techniques. The choice of LLMs aimed at ensuring a diverse representation of established models, considering the complexity of our benchmark tasks and the potential for erratic outputs from smaller LLMs when given the freedom to reason. For a comprehensive explanation of the querying process for these models, please refer to Appendix \ref{appendix:cotsb}.

Our investigation initially focused on probabilistic reasoning, which is a fundamental cognitive ability in decision-making. This ability facilitates the optimal integration of new information with pre-existing knowledge. We used the performance metric from the probabilistic reasoning experiment, namely posterior accuracy, which is calculated as one minus the deviation from the Bayes optimal prediction for each task. As depicted in Figure \ref{fig:cotsb}A, both CoT and SB techniques generally enhanced probabilistic reasoning compared to their base models, with CoT showing an average increase of $9.01\%$ and SB showing an increase of $3.10\%$.

Furthermore, we discovered that model-basedness, a critical aspect of reasoning and planning, is significantly augmented by both CoT and SB techniques, as shown in Figure \ref{fig:cotsb}B. Specifically, CoT demonstrated a $64.59\%$ increase, while SB showed a substantial increase of $118.59\%$.

Interestingly, a closer examination of the figures and the numerical data suggests that CoT is more effective for probabilistic reasoning, while SB excels in enhancing model-basedness. 
This observation aligns with the notion that step-by-step thinking can aid mathematical reasoning \cite{kojima2022large} while abstracting a problem by taking a step back can foster a better representation of the problem's abstract structure.
% Indeed, CoT was initially evaluated primarily for arithmetic and logic tasks . On the other hand, the strategy of ‘taking a step back’ in the ‘two-step-task’ emphasizes a fundamental factor of considering the previous step in the decision-making process, thereby eliciting model-based behavior. 
However, it is important to note that this analysis only serves as an initial observation. It does, nonetheless, highlight potential future applications of \texttt{CogBench} and illustrates how examining specific behaviors can provide valuable context, potentially guiding future decisions on the selection of one reasoning technique over another.

% \subsection{Behaviour 2: Optimism bias}
% Optimism bias is defined as the difference in learning rates between positive and negative prediction errors. This bias has been observed in humans \cite{lefebvre2017behavioural} and resource-rational agents \cite{schubert2023rational}.

% In our observations, LLMs exhibited a substantially higher level of optimism bias compared to humans. However, we hypothesized that by reducing resource constraints through the use of 'chain-of-thoughts' (CoT) and 'take-a-step-back' (SB) techniques, this bias in LLMs could be reduced closer to human bias.

% In our regression, we found that both CoT and SB techniques significantly reduced optimism bias, as indicated by the values $\beta=0.17\pm0.07,t=27.8,p<0.0001$ for CoT and $\beta=0.29\pm0.07,t=27.7,p<0.0001$ for SB. These techniques effectively helped LLMs to reduce their bias closer to the optimal human bias of $0.18$. Remarkably, the average optimism biases for CoT and SB were $0.14$ and $0.02$ respectively, lower than the human bias, and a significant improvement from the $0.31$ bias observed in LLMs prior to the application of these techniques.

% \\
% \\
% In summary, while there is room for further enhancement in performance, particularly in key areas, this section emphasizes that the majority of the modifications brought about by these techniques are behavioral. It demonstrates that even minor performance shifts resulting from a technique can be better comprehended by examining specific behavioral metrics that are well-established in the field of cognitive psychology.

\section{Discussion}

% Paragraph 1: summary of results

% Paragraph 2: state again benefits of cogsci approach and that benchmark is open-source

% Paragraph 3: mention the limits of transparency for proprietary models and "open-source models" for training data, to have better estimates

% Paragraph 4: how generalizable are results?

% % paragraph on controlled experiments?

% Paragraph 5: short conclusion

% % include potential future work? If so, what to mention?

% \\
% It helps cognitive science to have outside of distribution behaviors tested in their experiments which might put in question what metrics are being measured (e.g: learning rates show its limits?)
%restatable

% mention benchmark contributions as well
We have presented \texttt{CogBench}, a new open-source benchmark for evaluating LLMs. \texttt{CogBench} is rooted in well-established experimental paradigms from the cognitive psychology literature, providing a unique set of advantages over traditional LLM benchmarks. First, it is based on tried-and-tested experiments whose measures have been extensively validated over many years and shown to capture general cognitive constructs. In addition, unlike standard benchmarks, \texttt{CogBench} does not only focus on performance metrics alone but also comes with behavioral metrics that allow us to gain insights into \emph{how} a given task is solved. Finally, many of the included problems are procedurally-generated, thereby making it hard to game our benchmark by training on the test set. All our code and analysis will be publicly available, making it easy to use \texttt{CogBench} for the LLM community.

Our analyses yielded several key findings: as expected, RLHF enhanced the human-likeness of LLMs, while the number of parameters improved their performance and model-basedness. However, we also found surprising results. Despite expectations, code fine-tuning did not influence performance or model-basedness and open-source models exhibited less risk-taking behavior. Further, we found CoT prompting to be a promising choice for enhancing probabilistic reasoning. Conversely, SB prompting proved more effective for model-based reasoning.

While these results demonstrate the versatility of our benchmark, our analysis also faces several challenges. For instance, the limited transparency of certain proprietary models poses an issue to our regression analysis because acquiring details about certain models can be difficult or impossible. This lack of transparency could potentially affect the precision of our analysis. It also underscores the need for more transparency to facilitate more thorough and accurate evaluations \cite{laion2023, binz2023should}.

%The generalizability of our results is an important consideration. While our findings provide valuable insights into the behavior of LLMs, further research is needed to determine how these results generalize to other models or tasks.

% intersection also goes other way around
% can we design better experiments?
% example: exploration!
%Interestingly, the limitations in generalizability might present an opportunity for cognitive science. Testing out-of-distribution behaviors in their experiments, instead of human behaviors, might challenge our understanding of what metrics are being measured and lead to new insights.

Taken together, our study highlights the importance of behavioral metrics and cognitive modeling in evaluating LLMs and presents a novel benchmark for this purpose.
% Our findings underscore the role of model size and RLHF in improving LLM performance, model-basedness, and aligning it with human-like behavior. Furthermore, contrary to expectations, we found that training on code does not appear to impact performance and model-basedness and that proprietary models exhibit increased risk-taking behaviors. 
The analysis was preliminary and intended to provide a broad view of how \texttt{CogBench} can be used. The primary aim of this work is to equip the LLM community with new tools, inspired by cognitive science, to evaluate their models more comprehensively. Future work should focus on three areas. First, while cognitive science studies have demonstrated the external validity of the investigated tasks, it is yet to be shown for LLMs. Furthermore, we aim to extend the set of included tasks to cover a broader set of domains. Finally, we plan to properly automate our benchmark, mostly for prompt engineering techniques that were only briefly examined in this study. This could include studying the influence of impersonation \cite{salewski2023context}, meta-in-context learning \cite{coda2024meta}, and explanations \cite{lampinen2022can} on LLMs.
%Other prompt engineering, impersonation ,meta-in-context learning effects on behavior, in the cognitive  science field, the external validity of some of these measurement has been shown, and its unclear for LLMs and   should also be studies + add more tasks

%I should talk

% \section*{Acknowledgements}
% - Do not forget Sarah for the meta-cognition human data!!
% \textbf{Do not} include acknowledgements in the initial version of
% the paper submitted for blind review.

% If a paper is accepted, the final camera-ready version can (and
% probably should) include acknowledgements. In this case, please
% place such acknowledgements in an unnumbered section at the
% end of the paper. Typically, this will include thanks to reviewers
% who gave useful comments, to colleagues who contributed to the ideas,
% and to funding agencies and corporate sponsors that provided financial

\newpage
% \section*{Impacts statement}
% This paper presents work whose goal is to advance the field of Machine Learning and more specifically the evaluation of LLMs. There are many potential societal consequences of our work, none of which we feel must be specifically highlighted here.

\bibliography{main}

\begin{thebibliography}{61}
\providecommand{\natexlab}[1]{#1}
\providecommand{\url}[1]{\texttt{#1}}
\expandafter\ifx\csname urlstyle\endcsname\relax
  \providecommand{\doi}[1]{doi: #1}\else
  \providecommand{\doi}{doi: \begingroup \urlstyle{rm}\Url}\fi

\bibitem[Akata et~al.(2023)Akata, Schulz, Coda-Forno, Oh, Bethge, and Schulz]{akata2023playing}
Akata, E., Schulz, L., Coda-Forno, J., Oh, S.~J., Bethge, M., and Schulz, E.
\newblock Playing repeated games with large language models, 2023.

\bibitem[Almazrouei et~al.(2023)Almazrouei, Alobeidli, Alshamsi, Cappelli, Cojocaru, Debbah, Goffinet, Heslow, Launay, Malartic, Noune, Pannier, and Penedo]{falcon40b}
Almazrouei, E., Alobeidli, H., Alshamsi, A., Cappelli, A., Cojocaru, R., Debbah, M., Goffinet, E., Heslow, D., Launay, J., Malartic, Q., Noune, B., Pannier, B., and Penedo, G.
\newblock {Falcon-40B}: an open large language model with state-of-the-art performance.
\newblock 2023.

\bibitem[Anthropic(2023)]{claude2}
Anthropic.
\newblock Claude 2.
\newblock Blog post, 2023.
\newblock URL \url{https://www.anthropic.com/news/claude-2}.
\newblock Accessed: 2024-01-19.

\bibitem[Binz \& Schulz(2023)Binz and Schulz]{binz2022using}
Binz, M. and Schulz, E.
\newblock Using cognitive psychology to understand gpt-3.
\newblock \emph{Proceedings of the National Academy of Sciences}, 120\penalty0 (6):\penalty0 e2218523120, 2023.

\bibitem[Binz et~al.(2023)Binz, Alaniz, Roskies, Aczel, Bergstrom, Allen, Schad, Wulff, West, Zhang, et~al.]{binz2023should}
Binz, M., Alaniz, S., Roskies, A., Aczel, B., Bergstrom, C.~T., Allen, C., Schad, D., Wulff, D., West, J.~D., Zhang, Q., et~al.
\newblock How should the advent of large language models affect the practice of science?
\newblock \emph{arXiv preprint arXiv:2312.03759}, 2023.

\bibitem[Bommasani et~al.(2021)Bommasani, Hudson, Adeli, Altman, Arora, von Arx, Bernstein, Bohg, Bosselut, Brunskill, et~al.]{bommasani2021opportunities}
Bommasani, R., Hudson, D.~A., Adeli, E., Altman, R., Arora, S., von Arx, S., Bernstein, M.~S., Bohg, J., Bosselut, A., Brunskill, E., et~al.
\newblock On the opportunities and risks of foundation models.
\newblock \emph{arXiv preprint arXiv:2108.07258}, 2021.

\bibitem[Br{\"a}ndle et~al.(2021)Br{\"a}ndle, Binz, and Schulz]{brandle2021exploration}
Br{\"a}ndle, F., Binz, M., and Schulz, E.
\newblock Exploration beyond bandits.
\newblock \emph{The drive for knowledge: The science of human information seeking}, pp.\  147--168, 2021.

\bibitem[Brown et~al.(2020)Brown, Mann, Ryder, Subbiah, Kaplan, Dhariwal, Neelakantan, Shyam, Sastry, Askell, et~al.]{brown2020language}
Brown, T., Mann, B., Ryder, N., Subbiah, M., Kaplan, J.~D., Dhariwal, P., Neelakantan, A., Shyam, P., Sastry, G., Askell, A., et~al.
\newblock Language models are few-shot learners.
\newblock \emph{Advances in neural information processing systems}, 33:\penalty0 1877--1901, 2020.

\bibitem[Bubeck et~al.(2023)Bubeck, Chandrasekaran, Eldan, Gehrke, Horvitz, Kamar, Lee, Lee, Li, Lundberg, et~al.]{bubeck2023sparks}
Bubeck, S., Chandrasekaran, V., Eldan, R., Gehrke, J., Horvitz, E., Kamar, E., Lee, P., Lee, Y.~T., Li, Y., Lundberg, S., et~al.
\newblock Sparks of artificial general intelligence: Early experiments with gpt-4.
\newblock \emph{arXiv preprint arXiv:2303.12712}, 2023.

\bibitem[Burnell et~al.(2023)Burnell, Schellaert, Burden, Ullman, Martinez-Plumed, Tenenbaum, Rutar, Cheke, Sohl-Dickstein, Mitchell, et~al.]{burnell2023rethink}
Burnell, R., Schellaert, W., Burden, J., Ullman, T.~D., Martinez-Plumed, F., Tenenbaum, J.~B., Rutar, D., Cheke, L.~G., Sohl-Dickstein, J., Mitchell, M., et~al.
\newblock Rethink reporting of evaluation results in ai.
\newblock \emph{Science}, 380\penalty0 (6641):\penalty0 136--138, 2023.

\bibitem[Buschoff et~al.(2024)Buschoff, Akata, Bethge, and Schulz]{buschoff2024visual}
Buschoff, L. M.~S., Akata, E., Bethge, M., and Schulz, E.
\newblock Visual cognition in multimodal large language models, 2024.

\bibitem[Carpenter et~al.(2019)Carpenter, Sherman, Kievit, Seth, Lau, and Fleming]{Carpenter2019}
Carpenter, J., Sherman, M.~T., Kievit, R.~A., Seth, A.~K., Lau, H., and Fleming, S.~M.
\newblock Domain-general enhancements of metacognitive ability through adaptive training.
\newblock \emph{Journal of Experimental Psychology: General}, 148\penalty0 (1):\penalty0 51, 2019.

\bibitem[Cavagnaro et~al.(2016)Cavagnaro, Aranovich, McClure, Pitt, and Myung]{cavagnaro2016functional}
Cavagnaro, D.~R., Aranovich, G.~J., McClure, S.~M., Pitt, M.~A., and Myung, J.~I.
\newblock On the functional form of temporal discounting: An optimized adaptive test.
\newblock \emph{Journal of Risk and Uncertainty}, 52:\penalty0 233--254, 2016.

\bibitem[Chen et~al.(2021)Chen, Tworek, Jun, Yuan, de~Oliveira~Pinto, Kaplan, Edwards, Burda, Joseph, Brockman, Ray, Puri, Krueger, Petrov, Khlaaf, Sastry, Mishkin, Chan, Gray, Ryder, Pavlov, Power, Kaiser, Bavarian, Winter, Tillet, Such, Cummings, Plappert, Chantzis, Barnes, Herbert-Voss, Guss, Nichol, Paino, Tezak, Tang, Babuschkin, Balaji, Jain, Saunders, Hesse, Carr, Leike, Achiam, Misra, Morikawa, Radford, Knight, Brundage, Murati, Mayer, Welinder, McGrew, Amodei, McCandlish, Sutskever, and Zaremba]{humaneval}
Chen, M., Tworek, J., Jun, H., Yuan, Q., de~Oliveira~Pinto, H.~P., Kaplan, J., Edwards, H., Burda, Y., Joseph, N., Brockman, G., Ray, A., Puri, R., Krueger, G., Petrov, M., Khlaaf, H., Sastry, G., Mishkin, P., Chan, B., Gray, S., Ryder, N., Pavlov, M., Power, A., Kaiser, L., Bavarian, M., Winter, C., Tillet, P., Such, F.~P., Cummings, D., Plappert, M., Chantzis, F., Barnes, E., Herbert-Voss, A., Guss, W.~H., Nichol, A., Paino, A., Tezak, N., Tang, J., Babuschkin, I., Balaji, S., Jain, S., Saunders, W., Hesse, C., Carr, A.~N., Leike, J., Achiam, J., Misra, V., Morikawa, E., Radford, A., Knight, M., Brundage, M., Murati, M., Mayer, K., Welinder, P., McGrew, B., Amodei, D., McCandlish, S., Sutskever, I., and Zaremba, W.
\newblock Evaluating large language models trained on code, 2021.

\bibitem[Chen et~al.(2023)Chen, Liu, Shan, and Zhong]{doi:10.1073/pnas.2316205120}
Chen, Y., Liu, T.~X., Shan, Y., and Zhong, S.
\newblock The emergence of economic rationality of gpt.
\newblock \emph{Proceedings of the National Academy of Sciences}, 120\penalty0 (51):\penalty0 e2316205120, 2023.
\newblock \doi{10.1073/pnas.2316205120}.
\newblock URL \url{https://www.pnas.org/doi/abs/10.1073/pnas.2316205120}.

\bibitem[Christiano et~al.(2017)Christiano, Leike, Brown, Martic, Legg, and Amodei]{christiano2017deep}
Christiano, P.~F., Leike, J., Brown, T., Martic, M., Legg, S., and Amodei, D.
\newblock Deep reinforcement learning from human preferences.
\newblock \emph{Advances in neural information processing systems}, 30, 2017.

\bibitem[Cobbe et~al.(2021)Cobbe, Kosaraju, Bavarian, Chen, Jun, Kaiser, Plappert, Tworek, Hilton, Nakano, Hesse, and Schulman]{gsm8k}
Cobbe, K., Kosaraju, V., Bavarian, M., Chen, M., Jun, H., Kaiser, L., Plappert, M., Tworek, J., Hilton, J., Nakano, R., Hesse, C., and Schulman, J.
\newblock Training verifiers to solve math word problems, 2021.

\bibitem[Coda-Forno et~al.(2023)Coda-Forno, Witte, Jagadish, Binz, Akata, and Schulz]{coda2023inducing}
Coda-Forno, J., Witte, K., Jagadish, A.~K., Binz, M., Akata, Z., and Schulz, E.
\newblock Inducing anxiety in large language models increases exploration and bias.
\newblock \emph{arXiv preprint arXiv:2304.11111}, 2023.

\bibitem[Coda-Forno et~al.(2024)Coda-Forno, Binz, Akata, Botvinick, Wang, and Schulz]{coda2024meta}
Coda-Forno, J., Binz, M., Akata, Z., Botvinick, M., Wang, J., and Schulz, E.
\newblock Meta-in-context learning in large language models.
\newblock \emph{Advances in Neural Information Processing Systems}, 36, 2024.

\bibitem[Collins et~al.(2022)Collins, Wong, Feng, Wei, and Tenenbaum]{collins2022structured}
Collins, K.~M., Wong, C., Feng, J., Wei, M., and Tenenbaum, J.~B.
\newblock Structured, flexible, and robust: benchmarking and improving large language models towards more human-like behavior in out-of-distribution reasoning tasks.
\newblock \emph{arXiv preprint arXiv:2205.05718}, 2022.

\bibitem[Dasgupta et~al.(2020)Dasgupta, Schulz, Tenenbaum, and Gershman]{dasgupta2020theory}
Dasgupta, I., Schulz, E., Tenenbaum, J.~B., and Gershman, S.~J.
\newblock A theory of learning to infer.
\newblock \emph{Psychological review}, 127\penalty0 (3):\penalty0 412, 2020.

\bibitem[Dasgupta et~al.(2022)Dasgupta, Lampinen, Chan, Creswell, Kumaran, McClelland, and Hill]{dasgupta2022language}
Dasgupta, I., Lampinen, A.~K., Chan, S.~C., Creswell, A., Kumaran, D., McClelland, J.~L., and Hill, F.
\newblock Language models show human-like content effects on reasoning.
\newblock \emph{arXiv preprint arXiv:2207.07051}, 2022.

\bibitem[Daw et~al.(2011)Daw, Gershman, Seymour, Dayan, and Dolan]{daw2011model}
Daw, N.~D., Gershman, S.~J., Seymour, B., Dayan, P., and Dolan, R.~J.
\newblock Model-based influences on humans' choices and striatal prediction errors.
\newblock \emph{Neuron}, 69\penalty0 (6):\penalty0 1204--1215, 2011.

\bibitem[Ershadmanesh et~al.(2023)Ershadmanesh, Gholamzadeh, Desender, and Dayan]{Ershadmanesh2023}
Ershadmanesh, S., Gholamzadeh, A., Desender, K., and Dayan, P.
\newblock Meta-cognitive efficiency in learned value-based choice.
\newblock In \emph{2023 Conference on Cognitive Computational Neuroscience}, pp.\  29--32, 2023.
\newblock \doi{10.32470/CCN.2023.1570-0}.
\newblock URL \url{https://hdl.handle.net/21.11116/0000-000D-5BC7-D}.

\bibitem[Gershman(2018)]{gershman2018deconstructing}
Gershman, S.~J.
\newblock Deconstructing the human algorithms for exploration.
\newblock \emph{Cognition}, 173:\penalty0 34--42, 2018.

\bibitem[Google(2023)]{palm2bison}
Google.
\newblock Palm 2 technical report.
\newblock \emph{arXiv preprint arXiv:2305.10403}, 2023.

\bibitem[Hagendorff et~al.(2023)Hagendorff, Fabi, and Kosinski]{hagendorff2023human}
Hagendorff, T., Fabi, S., and Kosinski, M.
\newblock Human-like intuitive behavior and reasoning biases emerged in large language models but disappeared in chatgpt.
\newblock \emph{Nature Computational Science}, 3\penalty0 (10):\penalty0 833--838, 2023.

\bibitem[Hendrycks et~al.(2021)Hendrycks, Burns, Basart, Zou, Mazeika, Song, and Steinhardt]{mmlu}
Hendrycks, D., Burns, C., Basart, S., Zou, A., Mazeika, M., Song, D., and Steinhardt, J.
\newblock Measuring massive multitask language understanding, 2021.

\bibitem[Joshi et~al.(2017)Joshi, Choi, Weld, and Zettlemoyer]{triviaqa}
Joshi, M., Choi, E., Weld, D.~S., and Zettlemoyer, L.
\newblock Triviaqa: A large scale distantly supervised challenge dataset for reading comprehension, 2017.

\bibitem[Kaplan et~al.(2020)Kaplan, McCandlish, Henighan, Brown, Chess, Child, Gray, Radford, Wu, and Amodei]{kaplan2020scaling}
Kaplan, J., McCandlish, S., Henighan, T., Brown, T.~B., Chess, B., Child, R., Gray, S., Radford, A., Wu, J., and Amodei, D.
\newblock Scaling laws for neural language models.
\newblock \emph{arXiv preprint arXiv:2001.08361}, 2020.

\bibitem[Kojima et~al.(2022)Kojima, Gu, Reid, Matsuo, and Iwasawa]{kojima2022large}
Kojima, T., Gu, S.~S., Reid, M., Matsuo, Y., and Iwasawa, Y.
\newblock Large language models are zero-shot reasoners.
\newblock \emph{Advances in neural information processing systems}, 35:\penalty0 22199--22213, 2022.

\bibitem[LAION(2024)]{laion2023}
LAION.
\newblock Towards a transparent ai future: The call for less regulatory hurdles on open-source ai in europe.
\newblock Available at: \url{https://laion.ai/blog/transparent-ai/}, 2024.
\newblock Accessed: January 19, 2024.

\bibitem[Lampinen et~al.(2022)Lampinen, Dasgupta, Chan, Matthewson, Tessler, Creswell, McClelland, Wang, and Hill]{lampinen2022can}
Lampinen, A.~K., Dasgupta, I., Chan, S.~C., Matthewson, K., Tessler, M.~H., Creswell, A., McClelland, J.~L., Wang, J.~X., and Hill, F.
\newblock Can language models learn from explanations in context?
\newblock \emph{arXiv preprint arXiv:2204.02329}, 2022.

\bibitem[Lefebvre et~al.(2017)Lefebvre, Lebreton, Meyniel, Bourgeois-Gironde, and Palminteri]{lefebvre2017behavioural}
Lefebvre, G., Lebreton, M., Meyniel, F., Bourgeois-Gironde, S., and Palminteri, S.
\newblock Behavioural and neural characterization of optimistic reinforcement learning.
\newblock \emph{Nature Human Behaviour}, 1\penalty0 (4):\penalty0 0067, 2017.

\bibitem[Lejuez et~al.(2002)]{lejuez2002bart}
Lejuez, C.~W. et~al.
\newblock Evaluation of a behavioral measure of risk taking: the balloon analogue risk task (bart).
\newblock \emph{Journal of experimental psychology. Applied}, 8\penalty0 (2):\penalty0 75--84, 2002.
\newblock \doi{10.1037//1076-898x.8.2.75}.

\bibitem[Liu et~al.(2023)Liu, Wang, Sun, Yuan, Dong, Di, Wang, and Wang]{liu2023prompting}
Liu, X., Wang, J., Sun, J., Yuan, X., Dong, G., Di, P., Wang, W., and Wang, D.
\newblock Prompting frameworks for large language models: A survey, 2023.

\bibitem[Massey \& Wu(2005)Massey and Wu]{massey2005detecting}
Massey, C. and Wu, G.
\newblock Detecting regime shifts: The causes of under-and overreaction.
\newblock \emph{Management Science}, 51\penalty0 (6):\penalty0 932--947, 2005.

\bibitem[McCoy et~al.(2023)McCoy, Yao, Friedman, Hardy, and Griffiths]{mccoy2023embers}
McCoy, R.~T., Yao, S., Friedman, D., Hardy, M., and Griffiths, T.~L.
\newblock Embers of autoregression: Understanding large language models through the problem they are trained to solve.
\newblock \emph{arXiv preprint arXiv:2309.13638}, 2023.

\bibitem[McInnes et~al.(2020)McInnes, Healy, and Melville]{mcinnes2020umap}
McInnes, L., Healy, J., and Melville, J.
\newblock Umap: Uniform manifold approximation and projection for dimension reduction, 2020.

\bibitem[Montague et~al.(2012)Montague, Dolan, Friston, and Dayan]{montague2012computational}
Montague, P.~R., Dolan, R.~J., Friston, K.~J., and Dayan, P.
\newblock Computational psychiatry.
\newblock \emph{Trends in cognitive sciences}, 16\penalty0 (1):\penalty0 72--80, 2012.

\bibitem[MosaicML(2023)]{MosaicML2023Introducing}
MosaicML.
\newblock Introducing mpt-30b: Raising the bar for open-source foundation models.
\newblock Blog post, 2023.
\newblock URL \url{www.mosaicml.com/blog/mpt-30b}.
\newblock Accessed: 2023-06-22.

\bibitem[Nardo(2024)]{nardo2024}
Nardo, C.
\newblock The waluigi effect (mega-post).
\newblock Available at: \url{https://www.lesswrong.com/posts/D7PumeYTDPfBTp3i7/the-waluigi-effect-mega-post}, 2024.
\newblock Accessed: January 19, 2024.

\bibitem[OpenAI(2023)]{openai2023gpt4}
OpenAI.
\newblock Gpt-4 technical report.
\newblock \emph{arXiv preprint arXiv:2303.08774}, 2023.

\bibitem[Palminteri \& Lebreton(2022)Palminteri and Lebreton]{palminteri2022computational}
Palminteri, S. and Lebreton, M.
\newblock The computational roots of positivity and confirmation biases in reinforcement learning.
\newblock \emph{Trends in Cognitive Sciences}, 2022.

\bibitem[Patzelt et~al.(2018)Patzelt, Hartley, and Gershman]{patzelt2018computational}
Patzelt, E.~H., Hartley, C.~A., and Gershman, S.~J.
\newblock Computational phenotyping: using models to understand individual differences in personality, development, and mental illness.
\newblock \emph{Personality Neuroscience}, 1:\penalty0 e18, 2018.

\bibitem[Rescorla(1972)]{rescorla1972classical}
Rescorla, R.~A.
\newblock Classical conditioning ii: current research and theory.
\newblock pp.\ ~64, 1972.

\bibitem[Ruggeri et~al.(2022)Ruggeri, Panin, Vdovic, Ve{\'c}kalov, Abdul-Salaam, Achterberg, Akil, Amatya, Amatya, Andersen, et~al.]{ruggeri2022globalizability}
Ruggeri, K., Panin, A., Vdovic, M., Ve{\'c}kalov, B., Abdul-Salaam, N., Achterberg, J., Akil, C., Amatya, J., Amatya, K., Andersen, T.~L., et~al.
\newblock The globalizability of temporal discounting.
\newblock \emph{Nature Human Behaviour}, 6\penalty0 (10):\penalty0 1386--1397, 2022.

\bibitem[Salewski et~al.(2023)Salewski, Alaniz, Rio-Torto, Schulz, and Akata]{salewski2023context}
Salewski, L., Alaniz, S., Rio-Torto, I., Schulz, E., and Akata, Z.
\newblock In-context impersonation reveals large language models' strengths and biases.
\newblock \emph{arXiv preprint arXiv:2305.14930}, 2023.

\bibitem[Schaeffer et~al.(2023)Schaeffer, Miranda, and Koyejo]{schaeffer2023emergent}
Schaeffer, R., Miranda, B., and Koyejo, S.
\newblock Are emergent abilities of large language models a mirage?
\newblock \emph{arXiv preprint arXiv:2304.15004}, 2023.

\bibitem[Schurr et~al.(2023)Schurr, Reznik, Hillman, Bhui, and Gershman]{schurr2023dynamic}
Schurr, R., Reznik, D., Hillman, H., Bhui, R., and Gershman, S.~J.
\newblock Dynamic computational phenotyping of human cognition.
\newblock 2023.

\bibitem[Shekhar \& Rahnev(2021)Shekhar and Rahnev]{shekhar2021sources}
Shekhar, M. and Rahnev, D.
\newblock Sources of metacognitive inefficiency.
\newblock \emph{Trends in Cognitive Sciences}, 25\penalty0 (1):\penalty0 12--23, 2021.

\bibitem[Srivastava \& authors(2023)Srivastava and authors]{srivastava2023imitation}
Srivastava, A. and authors.
\newblock Beyond the imitation game: Quantifying and extrapolating the capabilities of language models, 2023.

\bibitem[Tamkin et~al.(2021)Tamkin, Brundage, Clark, and Ganguli]{tamkin2021understanding}
Tamkin, A., Brundage, M., Clark, J., and Ganguli, D.
\newblock Understanding the capabilities, limitations, and societal impact of large language models.
\newblock \emph{arXiv preprint arXiv:2102.02503}, 2021.

\bibitem[Touvron et~al.(2023)Touvron, Martin, Stone, Albert, Almahairi, Babaei, Bashlykov, Batra, Bhargava, Bhosale, et~al.]{touvron2023llama2}
Touvron, H., Martin, L., Stone, K., Albert, P., Almahairi, A., Babaei, Y., Bashlykov, N., Batra, S., Bhargava, P., Bhosale, S., et~al.
\newblock Llama 2: Open foundation and fine-tuned chat models.
\newblock \emph{arXiv preprint arXiv:2307.09288}, 2023.

\bibitem[Ullman(2023)]{ullman2023large}
Ullman, T.
\newblock Large language models fail on trivial alterations to theory-of-mind tasks.
\newblock \emph{arXiv preprint arXiv:2302.08399}, 2023.

\bibitem[Wei et~al.(2022)Wei, Tay, Bommasani, Raffel, Zoph, Borgeaud, Yogatama, Bosma, Zhou, Metzler, et~al.]{wei2022emergent}
Wei, J., Tay, Y., Bommasani, R., Raffel, C., Zoph, B., Borgeaud, S., Yogatama, D., Bosma, M., Zhou, D., Metzler, D., et~al.
\newblock Emergent abilities of large language models.
\newblock \emph{arXiv preprint arXiv:2206.07682}, 2022.

\bibitem[Wei et~al.(2023)Wei, Wang, Schuurmans, Bosma, Ichter, Xia, Chi, Le, and Zhou]{wei2023chainofthought}
Wei, J., Wang, X., Schuurmans, D., Bosma, M., Ichter, B., Xia, F., Chi, E., Le, Q., and Zhou, D.
\newblock Chain-of-thought prompting elicits reasoning in large language models, 2023.

\bibitem[Wilson et~al.(2014)Wilson, Geana, White, Ludvig, and Cohen]{wilson2014humans}
Wilson, R.~C., Geana, A., White, J.~M., Ludvig, E.~A., and Cohen, J.~D.
\newblock Humans use directed and random exploration to solve the explore--exploit dilemma.
\newblock \emph{Journal of Experimental Psychology: General}, 143\penalty0 (6):\penalty0 2074, 2014.

\bibitem[Yax et~al.(2023)Yax, Anll{\'o}, and Palminteri]{yax2023studying}
Yax, N., Anll{\'o}, H., and Palminteri, S.
\newblock Studying and improving reasoning in humans and machines.
\newblock \emph{arXiv preprint arXiv:2309.12485}, 2023.

\bibitem[Zheng et~al.(2023{\natexlab{a}})Zheng, Mishra, Chen, Cheng, Chi, Le, and Zhou]{zheng2023step}
Zheng, H.~S., Mishra, S., Chen, X., Cheng, H.-T., Chi, E.~H., Le, Q.~V., and Zhou, D.
\newblock Take a step back: Evoking reasoning via abstraction in large language models, 2023{\natexlab{a}}.

\bibitem[Zheng et~al.(2023{\natexlab{b}})Zheng, Chiang, Sheng, Zhuang, Wu, Zhuang, Lin, Li, Li, Xing, Zhang, Gonzalez, and Stoica]{zheng2023chatbotarena}
Zheng, L., Chiang, W.-L., Sheng, Y., Zhuang, S., Wu, Z., Zhuang, Y., Lin, Z., Li, Z., Li, D., Xing, E.~P., Zhang, H., Gonzalez, J.~E., and Stoica, I.
\newblock Judging llm-as-a-judge with mt-bench and chatbot arena.
\newblock \emph{arXiv preprint arXiv:2306.05685}, 2023{\natexlab{b}}.

\end{thebibliography}
\bibliographystyle{icml2024}

%%%%%%%%%%%%%%%%%%%%%%%%%%%%%%%%%%%%%%%%%%%%%%%%%%%%%%%%%%%%%%%%%%%%%%%%%%%%%%%
%%%%%%%%%%%%%%%%%%%%%%%%%%%%%%%%%%%%%%%%%%%%%%%%%%%%%%%%%%%%%%%%%%%%%%%%%%%%%%%
% APPENDIX
%%%%%%%%%%%%%%%%%%%%%%%%%%%%%%%%%%%%%%%%%%%%%%%%%%%%%%%%%%%%%%%%%%%%%%%%%%%%%%%
%%%%%%%%%%%%%%%%%%%%%%%%%%%%%%%%%%%%%%%%%%%%%%%%%%%%%%%%%%%%%%%%%%%%%%%%%%%%%%%
\newpage
\appendix
\onecolumn
\section{List of LLMs used}
\label{appendix:summaryofmodels}

% You can have as much text here as you want. The main body must be at most $8$ pages long.
% For the final version, one more page can be added.
% If you want, you can use an appendix like this one.  

% The $\mathtt{\backslash onecolumn}$ command above can be kept in place if you prefer a one-column appendix, or can be removed if you prefer a two-column appendix.  Apart from this possible change, the style (font size, spacing, margins, page numbering, etc.) should be kept the same as the main body.
%%%%%%%%%%%%%%%%%%%%%%%%%%%%%%%%%%%%%%%%%%%%%%%%%%%%%%%%%%%%%%%%%%%%%%%%%%%%%%%
%%%%%%%%%%%%%%%%%%%%%%%%%%%%%%%%%%%%%%%%%%%%%%%%%%%%%%%%%%%%%%%%%%%%%%%%%%%%%%%

{
\fontsize{5.5}{10}\selectfont  
\begin{longtable}{|l|r|l|l|l|r|r|l|l|}
  \hline
  Model Name & No. of Parameters & Finetuned LLM & Use of RLHF & Open Source & Size of Dataset & Context Length & Conversational & Code \\
  \hline
  \endhead
  GPT-4 & 1760 & No & Yes & No & 1.56 & 8 & No & No \\
  text-davinci-003 & 170 & No & Yes & No & 1.37 & 4 & No & No \\
  text-davinci-002 & 170 & No & No & No & 1.37 & 4 & No & No \\
  Claude-1 & 100 & No & RLAIF & No & 3.7 & 100 & No & No \\
  Claude-2 & 200 & No & RLAIF & No & 7.4 & 100 & No & No \\
  text-bison@002 & 340 & No & Yes & No & 1.4 & 8 & No & No \\
  Falcon-40b & 40 & No & No & Yes & 0.54 & 2 & No & No \\
  Falcon-40b-instruct & 40 & Falcon-40b & No & Yes & 0.6 & 2 & No & No \\
  MPT-30b & 30 & No & No & Yes & 1.76 & 8 & No & No \\
  MPT-30b-instruct & 30 & MPT-30b & No & Yes & 1.8 & 8 & No & No \\
  MPT-30b-chat & 30 & MPT-30b & No & Yes & 1.8 & 8 & Yes & No \\
  LLaMA-2-70 & 70 & No & No & Yes & 2 & 4 & No & No \\
  LLaMA-2-13 & 13 & No & No & Yes & 2 & 4 & No & No \\
  LLaMA-2-7 & 7 & No & No & Yes & 2 & 4 & No & No \\
  LLaMA-2-70-chat & 70 & Yes & Yes & Yes & 2 & 4 & Yes & No \\
  LLaMA-2-13-chat & 13 & Yes & Yes & Yes & 2 & 4 & Yes & No \\
  LLaMA-2-7-chat & 7 & Yes & Yes & Yes & 2 & 4 & Yes & No \\
  Vicuna-7b-v1.5 & 7 & LLaMA-2-7 & Yes & Yes & 2.37 & 4 & Yes & No \\
  Vicuna-13b-v1.5 & 13 & LLaMA-2-13 & Yes & Yes & 2.37 & 4 & Yes & No \\
  LLaMA-2-7b-longlora-100k-ft & 7 & LLaMA-2-7 & No & Yes & 2 & 100 & No & No \\
  LLaMA-2-7b-longlora-8k-ft & 7 & LLaMA-2-7 & No & Yes & 2 & 8 & No & No \\
  LLaMA-2-7b-longlora-16k-ft & 7 & LLaMA-2-7 & No & Yes & 2 & 16 & No & No \\
  LLaMA-2-7b-longlora-32k-ft & 7 & LLaMA-2-7 & No & Yes & 2 & 32 & No & No \\
  LLaMA-2-7b-longlora-32k & 7 & LLaMA-2-7 & No & Yes & 2 & 32 & No & No \\
  LLaMA-2-13b-longlora-32k-ft & 13 & LLaMA-2-13 & No & Yes & 2 & 32 & No & No \\
  LLaMA-2-13b-longlora-64k & 13 & LLaMA-2-13 & No & Yes & 2 & 64 & No & No \\
  LLaMA-2-13b-longlora-32k & 13 & LLaMA-2-13 & No & Yes & 2 & 32 & No & No \\
  LLaMA-2-70b-longlora-32k & 70 & LLaMA-2-70 & No & Yes & 2 & 32 & No & No \\
  LLaMA-2-70b-chat-longlora-32k & 70 & LLaMA-2-70-chat & Yes & Yes & 2 & 32 & Yes & No \\
  LongAlpaca-7B & 7 & LLaMA-2-7 & No & Yes & 2 & 16 & Yes & No \\
  LongAlpaca-13B & 13 & LLaMA-2-13 & No & Yes & 2 & 16 & Yes & No \\
  LongAlpaca-70B & 70 & LLaMA-2-70 & No & Yes & 2 & 16 & Yes & No \\
  CodeLlama-7B & 7 & LLaMA-2-7 & No & Yes & 2.5 & 16 & No & Yes \\
  CodeLlama-13B & 13 & LLaMA-2-13 & No & Yes & 2.5 & 16 & No & Yes \\
  CodeLlama-34B & 34 & LLaMA-2-34 & No & Yes & 2.5 & 16 & No & Yes \\
  \hline
  % \caption{Your table caption}
  % \label{your-label}
\end{longtable}
}
This table lists the 35 LLMs used in this paper with different features where: \begin{itemize}
\item \textbf{No. of Parameters:} This represents the number of parameters in the model, expressed in billions.
\item \textbf{Finetuned LLM:} This column indicates whether the model is a fine-tuned version of another model. If it is, the name of the original model from which it was fine-tuned is provided. If it is not a fine-tuned model, 'No' is written. However, if the model serves as the base model for another model listed in this table, 'Yes' is written.
\item \textbf{Use of RLHF:} This column specifies whether Reinforcement Learning from Human Feedback (RLHF) was used in the training of the model.
\item \textbf{Open Source:} This indicates whether the model is open source, meaning we have access to the weights of the model.
\item \textbf{Size of Dataset:} This represents the size of the dataset on which the model was trained, expressed in trillions of tokens.
\item \textbf{Context Length:} This refers to the length of the context available to the model during its operation.
\item \textbf{Conversational:} This indicates whether the model was fine-tuned with conversational datasets.
\item \textbf{Code:} This indicates whether the model was fine-tuned with code datasets.
\end{itemize}
Please note that the selection of features used for our analyses was made based on the best available knowledge of the authors, as some information about certain models can be challenging to obtain. This limitation could potentially impact the precision of the regression analysis. It underscores the need for greater transparency about LLMs to facilitate more thorough evaluations.

\newpage
\section{Comprehensive list \& explanation of the cognitive experiments}
\label{appendix:summaryoftasks}

\subsection{Probabilistic reasoning \cite{dasgupta2020theory} - Prior $\&$ likelihood weighting}

\subsubsection{Summary}
This experiment tests how agents update beliefs based on new evidence. Participants are given a wheel of fortune (representing initial prior probabilities) and two urns with different colored ball distributions (representing likelihoods). Upon drawing a ball, participants can revise their belief about the chosen urn, considering both the wheel (prior) and the ball color (evidence). The task allows testing adaptability to different prior/likelihood scenarios by changing the wheel division and ball distributions. Agents have to estimate the probability of the drawn ball’s urn. We use this task to estimate an agent's \textit{prior} and \textit{likelihood weightings}. In this task, people showed similar weighting between prior and likelihood, both under one. This underweighting is often referred to as system neglect \cite{massey2005detecting}.

\subsubsection{Methods}
We matched the probabilities used in \cite{dasgupta2020theory} to compare to human data. There they had either an informative likelihood case ($P(\text{left urn}|\text{red})=0.7, 0.8$ or $0.9$) and an informative prior ($P(\text{left urn}) = 0.5$ or $0.6$) or vice versa. They also trained humans on this experiment, so we only compared it to data from a human's first trial as we are not interested in learning but in how an LLM weighs its prior and likelihoods by default.
The default number of simulations here was 100. 

\subsubsection{Prompts for LLMs}
\begin{tcolorbox}[rounded corners, colback=mBlue!5!white,colframe=mBlue!75!black, width=\textwidth, title=\textbf{Example with informative likelihood}]

You are participating in an experiment where you are provided with a wheel of fortune and two urns. The wheel of fortune contains 10 evenly sized sections labeled either F or J, corresponding to the urns F and J. Another person will spin the wheel of fortune, select an urn based on the outcome of the spin, and then randomly pick a ball from the selected urn. Your goal is to give your best estimate of the probability of the urn being F after observing the ball drawn from the urn.
\\

Q: The wheel of fortune contains 6 sections labeled F and 4 sections labeled J. The urn F contains (8, 2) and the urn J contains (2, 8) red/blue balls. A red ball was drawn. What is the probability that it was drawn from Urn F? (Give your probability estimate on the scale from 0 to 1 rounded to two decimal places).
\\

A: I estimate the probability of the red ball to be drawn from the urn F to be 0.
\end{tcolorbox}

\subsubsection{Metrics}

\textbf{Performance:} Calculated as the posterior accuracy, therefore 1 minus the Bayes optimal.

\textbf{Behaviours 1 $\&$ 2: Prior and likelihood weightings}
A generalized version of Bayes rule considers prior $\beta_1$ and likelihood $\beta_2$ weightings to account for biases in Bayesian updating:

$$
P(A|B) \propto P(B|A)^{\beta_2} \cdot P(A)^{\beta_1}
$$

 For analytical convenience, this model can be reformulated as linear in log-odds. By fitting this model to the data using least squares linear regression, we can obtain the maximum likelihood estimates of the prior and likelihood weightings:

$$
\log \left( \frac{P(\text{Urn F} | \text{Ball})}{1 - P(\text{Urn F}| \text{Ball})} \right) = \beta_0 + \beta_1 \log \left( \frac{P(\text{Urn F})}{1 - P(\text{Urn F})} \right) + \beta_2 \log \left( \frac{P(\text{Ball} | \text{Urn F})}{P(\text{Ball} | \text{Urn J})} \right)
$$

- $P(\text{Urn F}| \text{Ball})$ is the subjective probability judgment of the urn being ‘F’ given the ball’s color.

- $P(\text{Urn F})$ and $P(\text{Ball} | \text{Urn F})$ are the prior probability and likelihood, respectively. 

- $\beta_1$ and $\beta_2$ are the prior and likelihood weightings, respectively, which are given as exponents in a generalized version of Bayes’ rule to capture specific biases. These two coefficients are the two behavioral metrics we report for this experiment.

- $\beta_0$ is the intercept term.

\subsection{Horizon task
\cite{wilson2014humans} - Directed $\&$ random exploration}
\subsubsection{Summary}
This task is a two-armed bandit task with stationary reward distributions. Agents first observe four reward values of randomly determined options, followed by making either one or six additional choices. We use this task to measure whether an agent uses uncertainty to guide its exploration behavior (\emph{directed exploration}) and/or whether it injects noise into its policy to explore (\emph{random exploration}). People are known to rely on a combination of both strategies when exploring \cite{wilson2014humans, brandle2021exploration}.

\subsubsection{Methods}
We followed the same methods for prompting LLMs as in \cite{binz2022using}. In the Horizon task, two distinct contexts are presented to participants, each differing in their time horizons. Each game involves 4 forced-choice trials, after which participants are given the opportunity to make a single choice (in the horizon 1 scenario) or six consecutive choices (in the horizon 6 scenario). The 4 forced-choice trials either offer one observation from one option and three from the other (unequal information condition), or two observations from each option (equal information condition).

The design of the horizon 1 and horizon 6 scenarios inherently provides a baseline for pure exploitation. Furthermore, the equal and unequal information conditions are designed to differentiate between directed and random exploration by examining the decision made in the first trial. In the equal information condition, a choice is categorized as random exploration if it aligns with the option with the lower average. Conversely, in the unequal information condition, a choice is classified as directed exploration if it aligns with the option that was observed less frequently during the forced-choice trials.

Our default number of simulations was 100.

\subsubsection{Prompts for LLMs}
\begin{tcolorbox}[rounded corners, colback=mBlue!5!white,colframe=mBlue!75!black, width=\textwidth, title=\textbf{Example with horizon 1 scenario}]
You are going to a casino that owns two slot machines. You earn money each time you play on one of these machines.
\\

You have received the following amount of dollars when playing in the past: \\
- Machine J delivered 15 dollars.\\
- Machine F delivered 37 dollars.\\
- Machine F delivered 28 dollars.\\
- Machine J delivered 11 dollars.\\
\\

Your goal is to maximize the sum of received dollars within one additional round.
\\

Q: Which machine do you choose?
\\

A: Machine

\end{tcolorbox}

\subsubsection{Metrics}

\textbf{Performance:} Average delivered dollars.
\\

\textbf{Behaviour 1 - Directed Exploration:} This metric is analyzed in the unequal information condition. Here, a regression is performed on the choice variable using three regressors:
\begin{itemize}
    \item x1 represents the difference in rewards,
    \item x2 represents the horizon (binary variable), and    
    \item x3 is the interaction term of x1 and x2 (i.e., $x1 \times x2$).
\end{itemize}

The beta coefficient for x2 (the presence or not of a horizon) is then extracted as the measure of directed exploration.

\textbf{Behaviour 2 - Random exploration:} We follow the same procedure as for the directed exploration but in the equal information condition to measure random exploration. However, in this case, the beta coefficient for x3 (the interaction effect between the difference in rewards and the presence of a horizon) from the regression provides the measure of random exploration.

\subsection{Restless bandit task \cite{Ershadmanesh2023} - Meta-cognition}

\subsubsection{Summary}
This is a two-armed bandit task with non-stationary reward distributions. There is always one option with a higher average reward. Every few trials a switch between the reward distributions of the two options occurs. Agents furthermore have to indicate after each choice how confident they are in their decisions. We use this task to measure \textit{meta-cognition}, which indicates whether an agent can assess the quality of its own cognitive abilities. People generally display this ability but its extent is influenced by various internal and external factors \cite{shekhar2021sources}. 

\subsubsection{Methods}
In each trial, LLMs are tasked with choosing between one arm which samples a reward from a normal distribution $N(60,8)$, while the other arm samples a reward from a $N(40,8)$. LLMs are informed that the slot machine with the higher average reward changes every 18-22 trials. 

Additionally, in each trial, LLMs must express their confidence in their choice on a scale from 0 to 1, as opposed to humans who use a Likert scale. The task is composed of 4 blocks, each containing 18-22 trials, resulting in approximately 80 trials in total. This is in contrast to the human task, which consists of 20 blocks for a total of 400 trials. The decision to limit the number of trials was made due to context size restrictions for some LLMs.

Our default number of simulations was 10.

\newpage

\subsubsection{Prompts for LLMs}
\begin{tcolorbox}[rounded corners, colback=mBlue!5!white,colframe=mBlue!75!black, width=\textwidth, title=\textbf{Example for reporting confidence at trial 23}]

Q: You are going to a casino that owns two slot machines named machine J and F. You earn dollars \$ each time you play on one of these machines with one machine always having a higher average \$ reward. Every 18 to 22 trials a switch of block takes place and the other slot machine will now give the higher point reward on average. However, you are not told about the change of block. After each choice, you have to indicate how confident you were about your choice being the best on a scale from 0 to 1. The casino includes 4 blocks of 18 to 22 trials, for a total of 80 trials 't'. Your goal is to interact with both machines and optimize your \$ as much as possible by identifying the best machine at a given point in time which comes in hand with being attentive to a potential change of block. The rewards will range between 20\$ and 80\$.                                     
\\

You have received the following amount of \$ when playing in the past: \\                                                
t=1: You chose J with a reported confidence of 0.43. It rewarded 54 \$.  \\                                              
t=2: You chose J with a reported confidence of 0.53. It rewarded 57 \$.    \\                                            
t=3: You chose J with a reported confidence of 0.88. It rewarded 70 \$.    \\                                            
...\\  

t=17: You chose F with a reported confidence of 0.99. It rewarded 59 \$. \\ 
t=18: You chose F with a reported confidence of 0.44. It rewarded 45 \$. \\                                              
t=19: You chose J with a reported confidence of 0.06. It rewarded 61 \$. \\
t=20: You chose J with a reported confidence of 0.51. It rewarded 64 \$. \\
t=21: You chose J with a reported confidence of 0.37. It rewarded 59 \$. \\
t=22: You chose J with a reported confidence of 0.54. It rewarded 42 \$. \\

Q: You are now in trial t=23. Which machine do you choose between machine J and F?(Think carefully remembering that exploration of both machines is required for optimal rewards. Give the answer in the form 'Machine $<$your choice$>$'.)
\\

A: Machine F.
\\

Q: How confident are you about your choice being the best on a continuous
scale running from 0 representing "
'this was a guess' to 1 representing 'very certain'? (Think carefully and give your answer to two decimal places)
\\

A: On a scale from 0 to 1, I am confident at 0.
\end{tcolorbox}

\subsubsection{Metrics}
\textbf{Performance:} Accuracy of choosing the best arm at a given trial.

\textbf{Behaviour - Meta-cognition:} We report the metacognitive sensitivity of a model by reporting the adjusted QSR \cite{Carpenter2019} defined as $$QSR = 1-(\text{accuracy} - \text{scaled confidence})^2$$ which is a standard metric for metacognitive sensitivity. The scaled confidence is computed as $$\text{scaled confidence} = \frac{\text{confidence} - \text{lowest reported confidence}}{\text{highest reported confidence} - \text{lowest reported confidence}}$$

\subsection{Experiment 2: Instrumental learning\cite{lefebvre2017behavioural} - Optimism bias \& learning rate}
\subsubsection{Summary}
\textbf{Instrumental learning} \cite{lefebvre2017behavioural}: LLMs encounter four two-armed bandit problems in an interleaved order. Each bandit problem is identified by a unique symbol pair. We use this task to investigate how an agent learns. First, we report the \textit{learning rate} of the agent which is common practice in two-armed bandits. Furthermore, we use it to reveal whether an agent learns more from positive than from negative prediction errors, i.e., whether it has an \emph{optimism bias}. People commonly display asymmetric tendencies when updating their beliefs by showing higher learning rates after encountering positive prediction errors compared to negative ones \cite{palminteri2022computational}.

\subsubsection{Methods}
As in \cite{lefebvre2017behavioural}, the task is 4 two-armed bandits of 96 trials (24 per slot machine). Here we randomly sample (without replacement) two letters for each to avoid biases towards a given letter. We used a cover story that involved a gambler visiting different casinos to generate our prompts. This choice has been inspired by similar tasks for human experiments \cite{gershman2018deconstructing}  and LLMs  \cite{binz2022using, coda2024meta}. Our default number of simulations per LLM is 10.

Casinos have the same reward probabilities as in the paper's first experiment: All arms have probabilities P=0.75 or 0.25 of winning 1 dollar and a reciprocal probability (1 – P) of getting nothing. In two casinos, the reward probability was the same for both arms (‘symmetric’ conditions), and in two other conditions, the reward probability was different across symbols (‘asymmetric’ conditions).

\subsubsection{Prompts for LLMs}
\begin{tcolorbox}[rounded corners, colback=mBlue!5!white,colframe=mBlue!75!black, width=\textwidth, title=\textbf{Example for 5th trial}]

You are going to visit four different casinos (named 1, 2, 3, and 4) 24 times each. Each casino owns two slot machines which all return either 1 or 0 dollars stochastically with different reward probabilities. Your goal is to maximize the sum of received dollars within 96 visits.\\

You have received the following amount of dollars when playing in the past: 

- Machine Q in Casino 4 delivered 0.0 dollars.\\
- Machine B in Casino 1 delivered 1.0 dollars.\\
- Machine B in Casino 1 delivered 0.0 dollars.\\
- Machine R in Casino 3 delivered 0.0 dollars.\\

Q: You are now in visit 5 playing in Casino 4. Which machine do you choose between Machine Q and Machine D? (Give the answer in the form "Machine $<$your choice$>$").\\

A: Machine
\end{tcolorbox}

\subsubsection{Metrics}
\textbf{Performance}: The performance is the average amount of money retrieved by the LLM.

\textbf{Behaviour 1 - Learning rate}:
We fit a Rescorla-Wagner model \cite{rescorla1972classical} which is standard to retrieve learning rates in two-armed bandits. This model operates under the assumption that decisions are made according to a Softmax function, which takes into account the predicted values of both arms. Each predicted value is updated using $\Delta V = \alpha \times$ prediction error where $\Delta V$ represents the change in value, and $\alpha$ denotes the learning rate. We report the learning rate which minimizes the negative log-likelihood.

\textbf{Behaviour 2 - Optimism bias}:
As in \cite{lefebvre2017behavioural}, we retrieve the optimism bias by assuming that there were two different learning rates, one for positive ($\alpha^+$) and one for negative ($\alpha^-$) prediction errors, sometimes called the RW $\pm$ model. The two learning rates were fit in the same way as for the standard Rescorla-Wagner model and the Optimism bias is computed as $\alpha^+ - \alpha^-$. This measure provides a quantitative representation of an individual’s tendency to learn more from positive outcomes than from negative ones.

\subsection{Two step task \cite{daw2011model} - Model-basedness}

\subsubsection{Summary}
This is a decision-making task in which agents have to accumulate as many treasures as possible. Taking an action from a starting state transfers the agent to one out of two second-stage states. In each of these second-stage states, the agent has the choice between two options that probabilistically lead to treasures. Finally, the agent is transferred back to the initial state and the process repeats for a predefined number of rounds. The task experimentally disentangles model-based from model-free reinforcement learning. We therefore use it to measure an agent's \emph{model-basedness}. Previous studies using this task have shown that people rely on a combination of model-free and model-based reinforcement learning \cite{daw2011model}.

\subsubsection{Methods}
We followed the same methods for LLMs as in \cite{binz2022using} with a 20-day horizon. Our default number of simulations was 100.

The transition probabilities from the first stage to the chosen second stage are fixed at 70$\%$. The two-step task gauges model-based decision-making by observing how past outcomes influence current choices. If a participant’s decisions reflect the previous trial’s second-stage state and reward, it suggests model-based decision-making, as they’re using a cognitive model of the task. However, if decisions are solely based on the previous trial’s first-stage choice and reward, it indicates model-free decision-making.

\subsubsection{Prompts for LLMs}
\begin{tcolorbox}[rounded corners, colback=mBlue!5!white,colframe=mBlue!75!black, width=\textwidth, title=\textbf{Example on 5th day after choosing planet Y for the first-step of the task.}]

You will travel to foreign planets in search of treasures.
When you visit a planet, you can choose an alien to trade with. The chance of getting treasures from these aliens changes over time. Your goal is to maximize the number of received treasures.
\\

Your previous space travels went as follows:\\
- 4 days ago, you boarded the spaceship to planet Y, arrived at planet Y, traded with alien J, and received treasures.\\
- 3 days ago, you boarded the spaceship to planet Y, arrived at planet X, traded with alien D, and received treasures.\\
- 2 days ago, you boarded the spaceship to planet Y, arrived at planet Y, traded with alien J, and received junk. \\
- 1 day ago, you boarded the spaceship to planet Y, arrived at planet X, traded with alien D, and received treasures.\\

Q: Do you want to take the spaceship to planet X or planet Y?\\

A: Planet Y.
\\
You arrive at planet Y.
\\

Q: Do you want to trade with alien J or K?
\\

A: Alien
\end{tcolorbox}

\subsubsection{Metrics}
\textbf{Performance:} Average number of received treasures. It is worth noting that the design of this experiment was done in a way that being model-free or model-based retrieves the same amount of rewards in average.
\\

\textbf{Behaviour - Model-basedness:}
To retrieve the model-basedness of an agent, we compute a regression using three regressors:
\begin{itemize}
    \item x1 representing rewards,
    \item x2 representing common transitions (binary variable) and
    \item x3 is the interaction term of x1 and x2 (i.e., $x1 \times x2$).
\end{itemize}
The regression is performed with the ‘stay probabilities’ as the dependent variable, and x1, x2, and x3 as the independent variables. The ‘stay probabilities’ represent the likelihood of a participant repeating the same first-stage choice on the next trial. We then retrieve the beta parameter for the interaction effect.

In essence, the interaction effect captures how the influence of rewards on stay probabilities changes depending on whether the previous trial involved a common or rare transition. A significant beta parameter for x3 would suggest that the effect of rewards on stay probabilities is not the same for common and rare transitions, indicating the presence of model-based decision-making.

\subsection{Temporal discounting \cite{ruggeri2022globalizability}}
\subsubsection{Summary}
Agents have to make a series of choices between two options. Each option is characterized by a monetary outcome and an associated delay until the outcome is received. We use this task to assess \emph{temporal discounting}, indicating whether an agent prefers smaller but immediate gains over larger delayed ones. People generally show a preference for immediate gains, although the precise functional form of their discounting is still a matter of debate \cite{cavagnaro2016functional}.

\subsubsection{Methods}
This task tests discounting patterns from three baseline scenarios to determine preference for immediate or delayed choices for gains (at two magnitudes) and losses (one). Second, they analyzed the prevalence of all choice anomalies using 4 additional items. Participants responded to 10 to 13 questions, depending on their responses to the initial three sets. Each baseline consisted of five sub-questions. Individuals saw at most three sub-questions depending on the order of their choices. 
It is worth noting that since this task is the only one which is not procedurally generated, there is only one simulation needed per LLM.
\subsubsection{Prompts for LLMs}

\begin{tcolorbox}[rounded corners, colback=mBlue!5!white,colframe=mBlue!75!black, width=\textwidth, title=\textbf{Examples for first baseline}]

Q: What do you prefer between the following two options:\\
 - Option 1: Receive 500 dollars now.\\
 - Option 2: Receive 550 dollars in 12 months.\\
A: I prefer option 2.\\

Q: What do you prefer between the following two options:\\
 - Option 1: Receive 500 dollars now.\\
 - Option 2: Receive 600 dollars in 12 months.\\
 
A: I prefer option

\end{tcolorbox}

\begin{tcolorbox}[rounded corners, colback=mBlue!5!white,colframe=mBlue!75!black, width=\textwidth, title=\textbf{Examples for 2nd baseline (different magnitude)}]

Q: What do you prefer between the following two options:\\
 - Option 1: Receive 5000 dollars now.\\
 - Option 2: Receive 5500 dollars in 12 months.\\
A: I prefer option 1.\\

Q: What do you prefer between the following two options:\\
 - Option 1: Receive 5000 dollars now.\\
 - Option 2: Receive 5100 dollars in 12 months.\\
A: I prefer option 1.\\

“Q: What do you prefer between the following two options:\\
 - Option 1: Receive 5000 dollars now.\\
 - Option 2: Receive 5050 dollars in 12 months.\\
 
A: I prefer option

\end{tcolorbox}

\begin{tcolorbox}[rounded corners, colback=mBlue!5!white,colframe=mBlue!75!black, width=\textwidth, title=\textbf{Examples for 3rd baseline (loss as opposed to gain)}]

Q: What do you prefer between the following two options:\\
 - Option 1: Pay 500 dollars now.\\
 - Option 2: Pay 550 dollars in 12 months.\\
 
A: I prefer option 1\\

Q: What do you prefer between the following two options:\\
 - Option 1: Pay 500 dollars now.\\
 - Option 2: Pay 510 dollars in 12 months.\\
 
A: I prefer option 1\\

Q: What do you prefer between the following two options:        \\                        
 - Option 1: Pay 500 dollars now.      \\                                                
 - Option 2: Pay 505 dollars in 12 months. \\                                  
 
A: I prefer option

\end{tcolorbox}

\begin{tcolorbox}[rounded corners, colback=mBlue!5!white,colframe=mBlue!75!black, width=\textwidth, title=\textbf{Example for testing present bias}]

Q: What do you prefer between the following two options:\\
 - Option 1: Receive 500 dollars in 12 months.\\
 - Option 2: Receive 600 dollars in 24 months.\\
 
A: I prefer option

\end{tcolorbox}

\begin{tcolorbox}[rounded corners, colback=mBlue!5!white,colframe=mBlue!75!black, width=\textwidth, title=\textbf{Example for testing subbaddictivity}]

Q: What do you prefer between the following two options:\\
 - Option 1: Receive 500 dollars now.\\
 - Option 2: Receive 700 dollars in 24 months.\\
 
A: I prefer option

\end{tcolorbox}

\begin{tcolorbox}[rounded corners, colback=mBlue!5!white,colframe=mBlue!75!black, width=\textwidth, title=\textbf{Example for testing delay-speedup asymmetry}]

Q: What do you prefer between the following two options:\\
 - Option 1: Receive 500 dollars now.\\
 - Option 2: Wait 12 months for the 500 dollars but with an additional 99 dollars.\\
 
A: I prefer option

\end{tcolorbox}

\begin{tcolorbox}[rounded corners, colback=mBlue!5!white,colframe=mBlue!75!black, width=\textwidth, title=\textbf{Example for testing delay-length asymmetry}]

Q: What do you prefer between the following two options:\\
 - Option 1: Wait 12 months to receive 600 dollars now.\\
 - Option 2: Pay 100 dollars and receive the 600 dollars gain now.\\
 
A: I prefer option

\end{tcolorbox}

\subsubsection{Metrics}
\textbf{Behaviour:} We only report one temporal discounting metric as behavioral. We use the same scoring system as in \cite{ruggeri2022globalizability} where the score is between 0 and 19. \\

The constructed score corresponded to the ranking of the sub-question where the individual switched from choosing the delayed option towards the sooner option. An individual who always chose the latter option was assigned a score of 0. An individual who always chose the sooner option was assigned a score of 5 (for baselines) and 1 for each of the 4 biases. Adding this score across questions results in a total score of 0 for individuals always choosing the delayed amount and 19 (5+5+5+4) for individuals always choosing the sooner amount. 

\subsection{Balloon Analogue Risk Task (BART) 
 \cite{lejuez2002bart} - Risk}
\subsubsection{Summary}
Agents have to inflate an imaginary balloon to obtain rewards. They may choose to stop inflating and cashing out all rewards accumulated so far. There is a chance that the balloon pops at any point in time and all rewards will be lost. We use this task to assess \emph{risk-taking} behavior. Human risk-taking in this task is \say{significantly correlated with scores on self-report  measures  of  risk-related  constructs  and  with  the  self-reported occurrence of real-world risk behaviors} \cite{lejuez2002bart}.

\subsubsection{Methods}
We substitute the colors of the balloons with the letters A, B, and C. We reduce the number of balloons from 15 to 10 per category to avoid context-size limits from some LLMs. Our default number of simulations per LLM is 10.

For the three balloons we sample from the same three distributions as in the paper where probabilities increase when inflating the balloon but start at $1/32, 1/8$ and $1/128$ per different balloon category. We randomly shuffle the assignment to A, B, and C to avoid biases of LLMs towards certain letters.

\subsubsection{Prompts for LLMs}
\begin{tcolorbox}[rounded corners, colback=mBlue!5!white,colframe=mBlue!75!black, width=\textwidth, title=\textbf{Example for 5th balloon}]

In this game, you will encounter 3 different balloons labeled A, B, and C. There will be a total of 10 balloons for each type of balloon. Your goal is to accumulate as many points as possible without popping the balloon. You will be presented with a balloon and given the option to inflate it or not. Each inflation increases the balloon's size and potential points but also carries a risk of the balloon popping. Your task is to decide whether to inflate the balloon or not knowing that a successful inflation adds 1 point from that balloon. Once you decide to stop inflating the balloon, you can no longer earn points from that balloon. If the balloon pops before you stop inflating, you will lose all the points accumulated in that balloon. Your final score will be determined by the total number of points earned across all 30 balloons. Your goal is to maximize your final score. \\

 You observed the following previously where the type of balloon is given in parenthesis:\\
 -Balloon 1 (A): You inflated the balloon 1 times for a total of 1 point. It did not explode.\\
 -Balloon 2 (C): You inflated the balloon 4 times for a total of 4 points. It did not explode.\\
 -Balloon 3 (A): You inflated the balloon 7 times for a total of 0 points. It did explode.\\
 -Balloon 4 (C): You inflated the balloon 5 times for a total of 5 points. It did not explode.\\
 -Balloon 5 (A): You inflated the balloon 9 times for a total of 0 points. It did explode.
\\

Q: You are currently with Balloon 5 which is a balloon of type A. What do you do? (Option 1 for 'skip' or 0 for 'inflate')
\\

A: Option 
\end{tcolorbox}

\subsubsection{Metrics}
\textbf{Performance:} The performance is the average points across all simulations.

\textbf{Behaviour: Risk} In the paper they report the adjusted risk which is defined as the average number of pumps excluding balloons that exploded. However, this does not take into account edge behaviours which always inflate which is the case for some LLMs and therefore we decided to report the risk as the average number of inflation attempts.

\newpage
\section{Full benchmark results for rest of LLMs}
\label{appendix:allresults}
\begin{figure*}[!htbp]
\begin{tabular}{l}
    \includegraphics[width=0.9\textwidth]{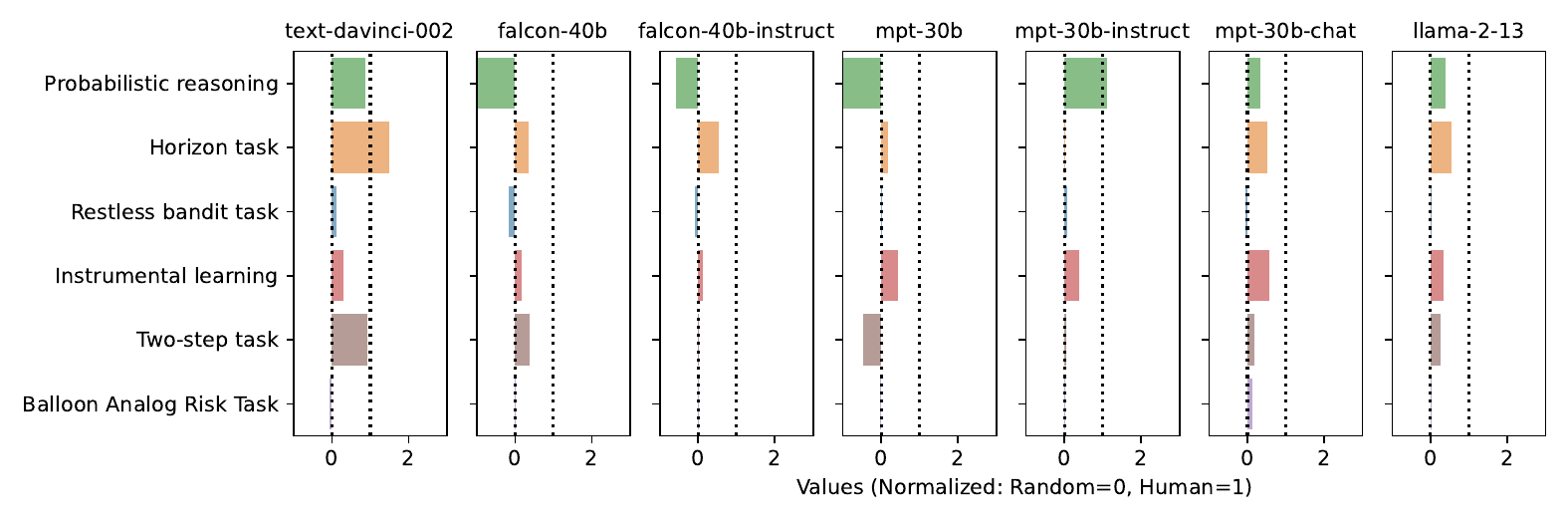}
    \\
    \includegraphics[width=0.9\textwidth]{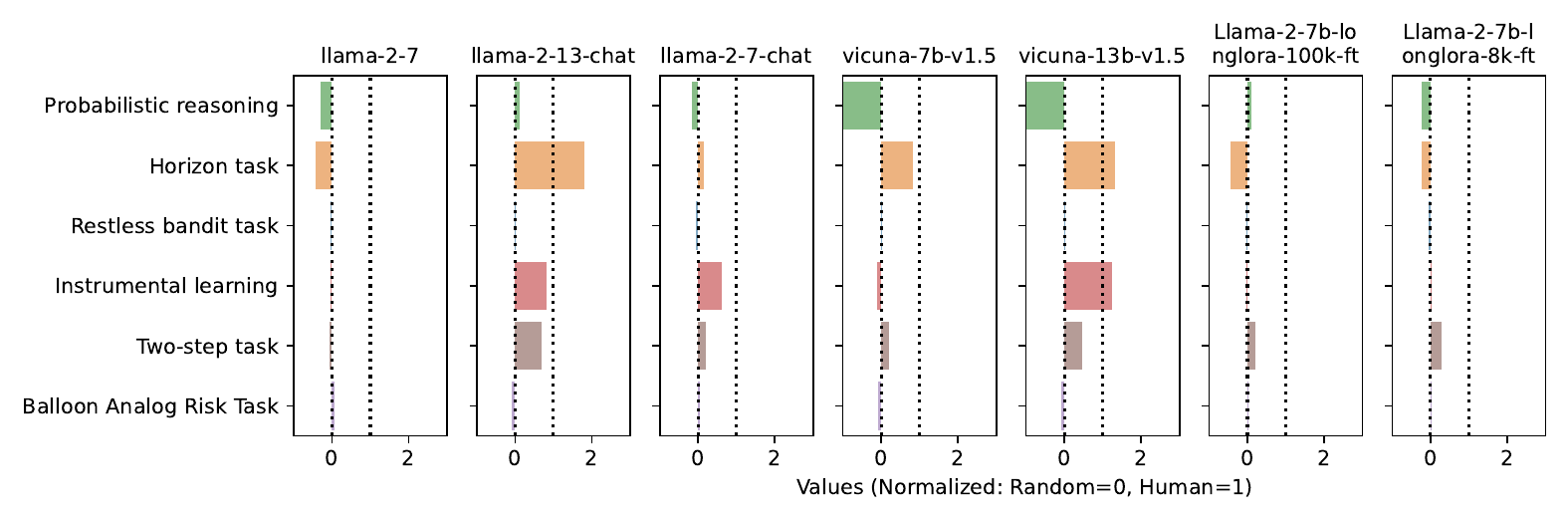}
    \\
    \includegraphics[width=0.9\textwidth]{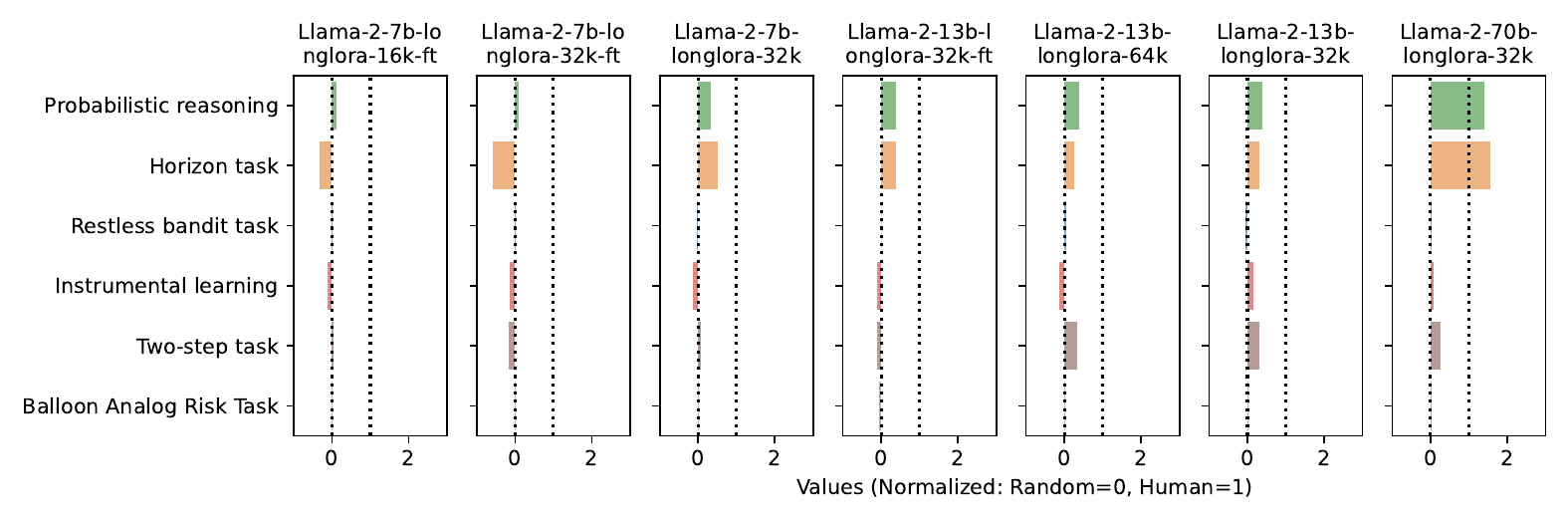}
    \\
    \includegraphics[width=0.9\textwidth]{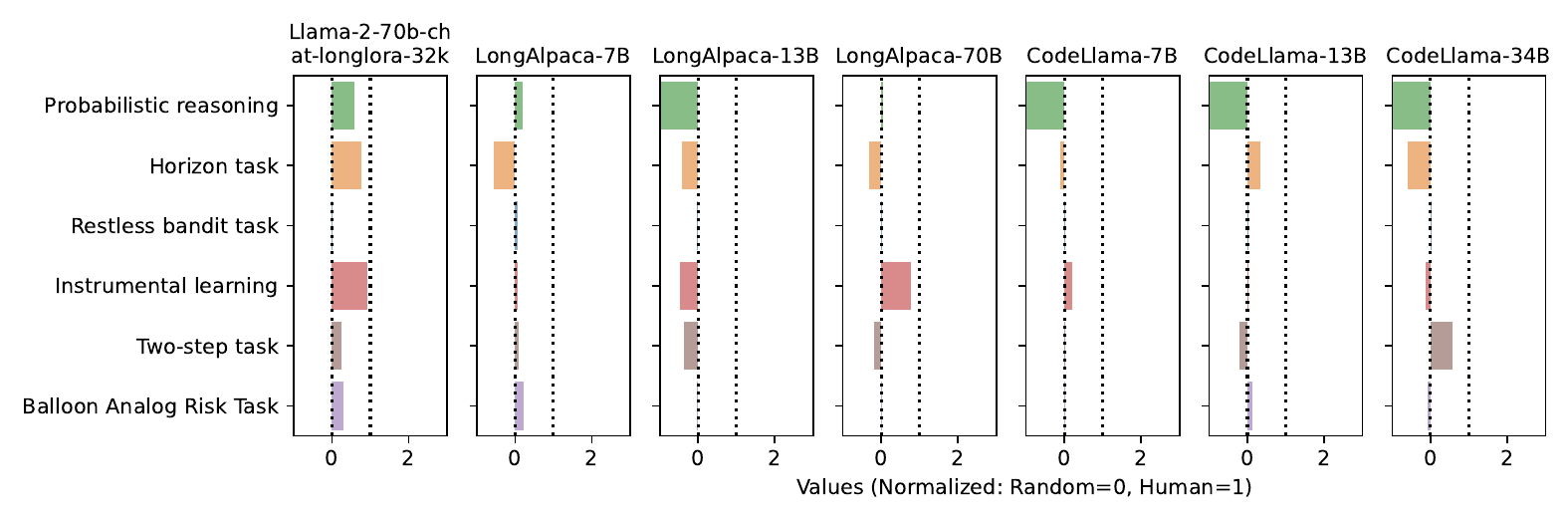}
\end{tabular}
\caption{Performance metrics}
\end{figure*}

\begin{figure*}[!htbp]
\begin{tabular}{l}
    \includegraphics[width=0.9\textwidth]{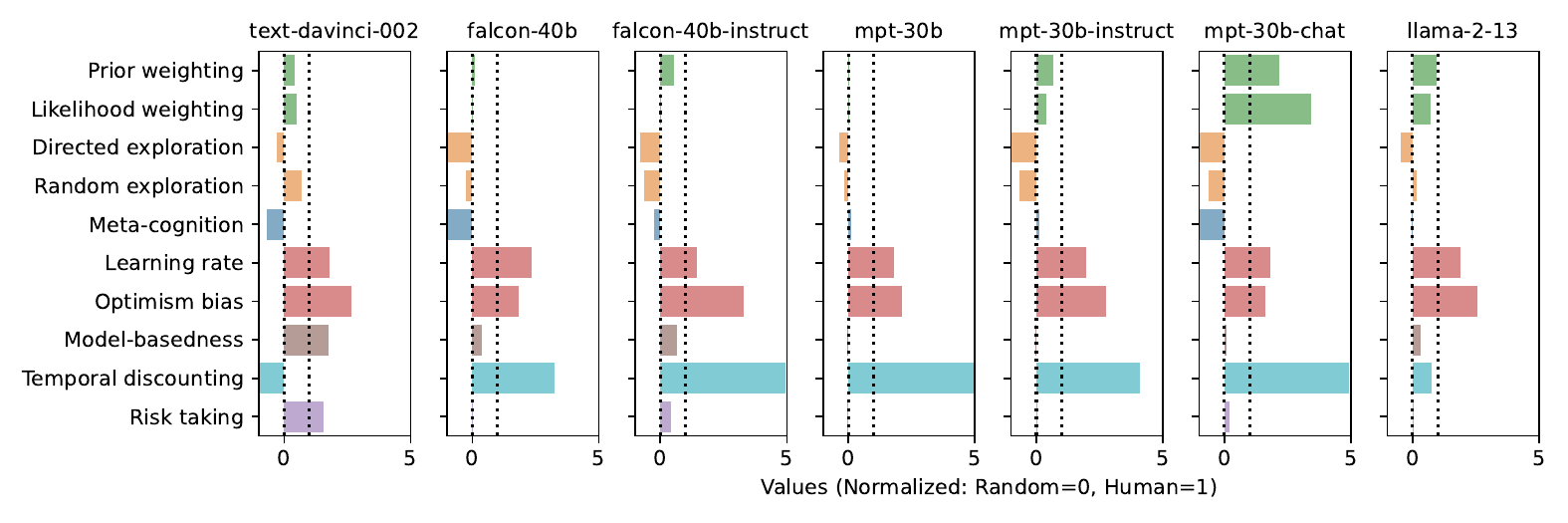}
    \\
    \includegraphics[width=0.9\textwidth]{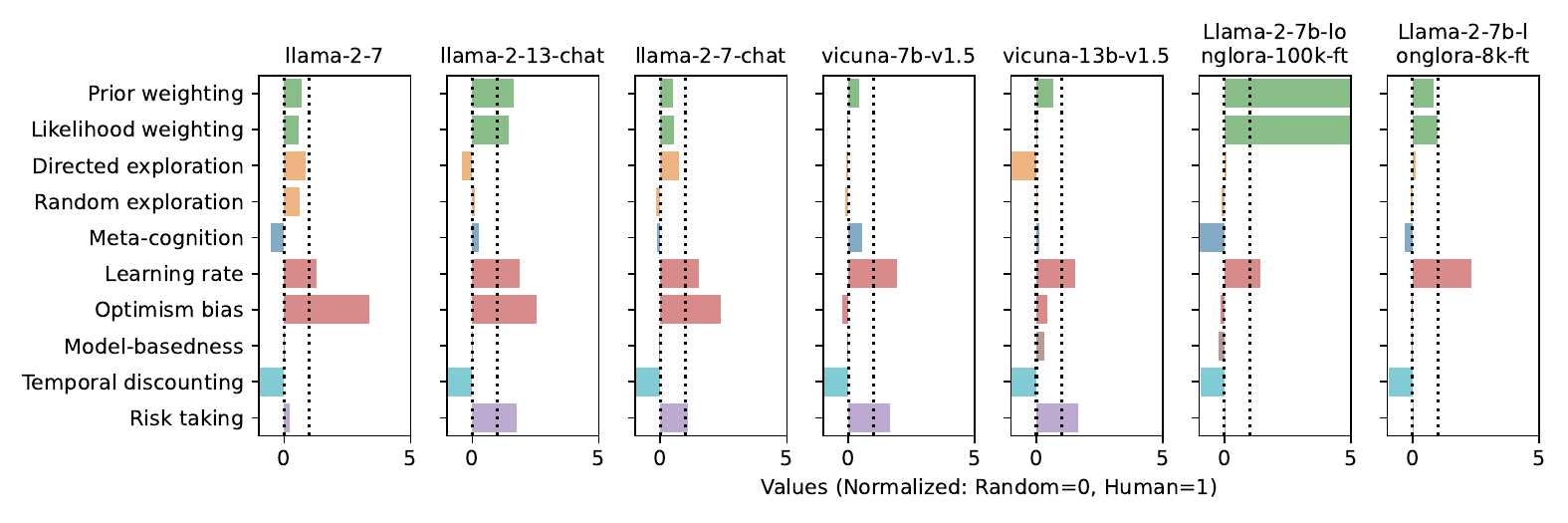}
    \\
    \includegraphics[width=0.9\textwidth]{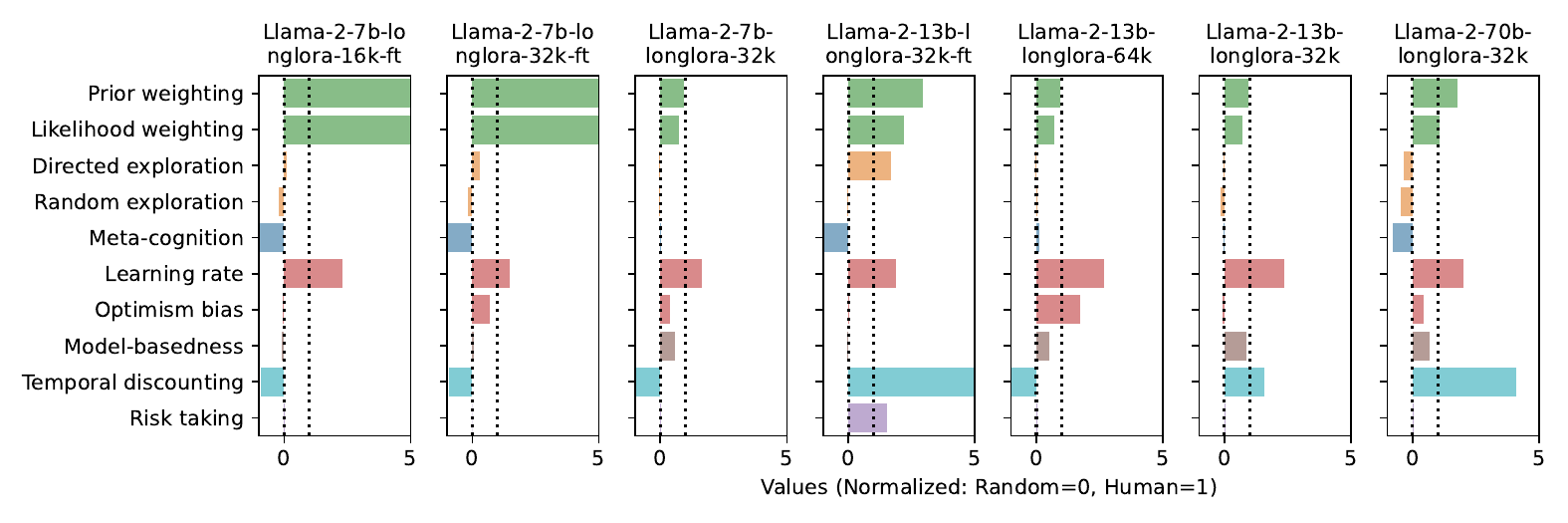}
    \\
    \includegraphics[width=0.9\textwidth]{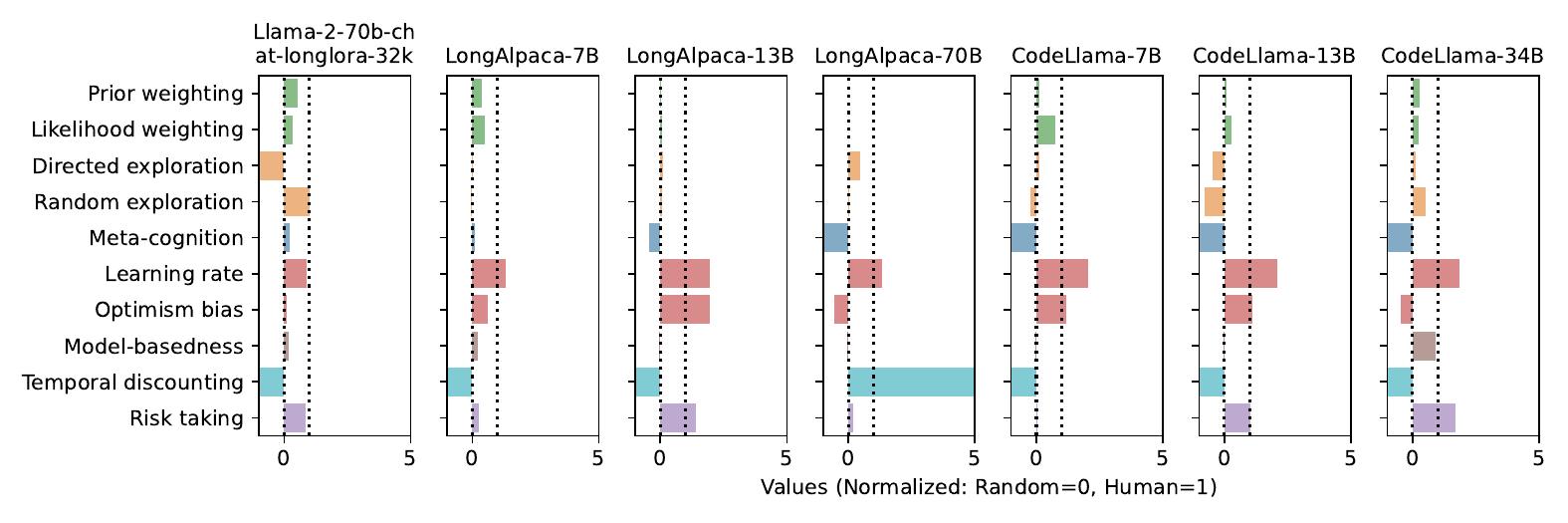}
\end{tabular}
\caption{Behavioral metrics}
\end{figure*}

\newpage

\section{Prompt Engineering techniques}
\label{appendix:cotsb}

In both the CoT and SB experiments, we appended specific prompts at the end (details provided below) where the function ‘self.format\_answer’ was different for each experiment. We imposed a limit of 300 tokens for an LLM. This approach, however, presented some challenges when compared with the standard benchmark analysis, which is designed to output a maximum of one token to ensure a context that enforces a one-token answer.

When we permit an LLM to modify the context with a flexible number of tokens, despite our attempts to enforce a maximum word limit, some LLMs do not consistently adhere to this constraint. This flexibility introduces complexity into the process of automating these engineering techniques across different experiments for various types of LLMs.

Additionally, some LLMs begin to exhibit chaotic behavior, and once this occurs, it becomes difficult to revert to a controlled state. This phenomenon, known as the ‘Waluigi effect’ \cite{nardo2024}, underscores the challenges of managing the balance between flexibility and control in the design and operation of LLMs. 

\begin{tcolorbox}[rounded corners, colback=mBlue!5!white,colframe=mBlue!75!black, width=\textwidth, title=\textbf{Example for take-a-step-back}]
First, take-a-step-back and think in the following two steps to answer this:\\
Step 1) Abstract the key concepts and principles relevant to this question in a maximum of 60 words."\\
Step 2) Use the abstractions to reason through the question in a maximum of 60 words.\\

Finally, give your final answer in the format 'Final answer: \{self.format\_answer\}$<$your choice$>$'. It is very important that you always answer in the right format even if you have no idea or you believe there is not enough information.\\

A: Step 1)
\end{tcolorbox}

\begin{tcolorbox}[rounded corners, colback=mBlue!5!white,colframe=mBlue!75!black, width=\textwidth, title=\textbf{Example for chain-of-thought}]

First break down the problem into smaller steps and reason through each step logically in a maximum of 100 words before giving your final answer in the format 'Final answer: \{self.format\_answer\}$<$your choice$>$'. It is very important that you always answer in the right format even if you have no idea or you believe there is not enough information.
\\

A: Let's think step by step:

\end{tcolorbox}

% % \label{appendix:cotsb}
% \begin{figure*}[h]
% \begin{tabular}{lll}
%         \textbf{A} & \hspace{0cm} \textbf{B} & \textbf{C} \\
%         \includegraphics[width=0.3\textwidth]{figures/cot_sb_small/posterior.pdf} &
%         \includegraphics[width=0.3\textwidth]{figures/cot_sb_small/modelbasedness.pdf} &
%         \includegraphics[width=0.3\textwidth]{figures/cot_sb_small/optimismbias.pdf}    
% \end{tabular}
% \caption{CoT, SB and base model results on \textbf{A:} Posterior accuracy, \textbf{B:} Model-basedness and \textbf{C:} Optimism bias}
% \label{fig:appendix_cotsb}

% \end{figure*}

% \begin{figure*}[t]
% \begin{tabular}{ll}
%         \textbf{A} & \hspace{0cm} \textbf{B} \\
%         \includegraphics[\textwidth]{figures/cot_sb/posterior.pdf} &
%         \includegraphics[\textwidth]{figures/cot_sb/modelbasedness.pdf} \\
%         \textbf{C} & \\
%         \includegraphics[\textwidth]{figures/cot_sb/optimismbias.pdf}    
% \end{tabular}
% \end{figure*}

\end{document}